\documentclass{article}

\usepackage[final,nonatbib]{neurips_2021}

\usepackage[numbers]{natbib}
\usepackage[utf8]{inputenc} \usepackage[T1]{fontenc}    \usepackage{hyperref}       \usepackage{url}            \usepackage{booktabs}       \usepackage{amsfonts}       \usepackage{nicefrac}       \usepackage{xcolor}
\usepackage{epsfig}
\usepackage{algorithm}
\usepackage{algorithmic}
\usepackage{amsthm}
\usepackage{amsmath}
\usepackage{microtype}
\usepackage{mathtools}
\usepackage{float}
\usepackage{graphicx}
\usepackage{wrapfig}
\usepackage{makecell}
\usepackage{caption}
\usepackage{subcaption}
\usepackage{bm}
\usepackage{bbm}
\usepackage{multirow}
\usepackage{microtype}
\usepackage{enumitem}
\usepackage{cancel}
\usepackage{chngcntr}

\newcommand{\one}[1]{\mathbbm{1}_{[#1]}}

\title{Intriguing Properties of Contrastive Losses}

\author{Ting Chen\\
  Google Research\\
  {\tt\small iamtingchen@google.com}\\
  \And 
  Calvin Luo\\
  Google Research\\
  {\tt\small calvinluo@google.com}
  \And 
  Lala Li\\
  Google Research\\
  {\tt\small lala@google.com}
}

\renewcommand\footnotemark{}

\begin{document}
\maketitle

\begin{abstract}
We study three intriguing properties of contrastive learning. 
First, we generalize the standard contrastive loss to a broader family of losses, and we find that various instantiations of the generalized loss perform similarly under the presence of a multi-layer non-linear projection head.
Second, we study if instance-based contrastive learning (with a global image representation) can learn well on images with multiple objects present. We find that meaningful hierarchical local features can be learned despite the fact that these objectives operate on global instance-level features. 
Finally, we study the phenomenon of \textit{feature suppression} among competing features shared across augmented views, such as ``color distribution'' vs ``object class''. We construct datasets with explicit and controllable competing features, and show that, for contrastive learning, a few bits of easy-to-learn shared features can suppress, and even fully prevent, the learning of other sets of competing features.
In scenarios where there are multiple objects in an image, the dominant object would suppress the learning of smaller objects.
Existing contrastive learning methods critically rely on data augmentation to favor certain sets of features over others, and could suffer from learning saturation for scenarios where existing augmentations cannot fully address the feature suppression.
This poses open challenges to existing contrastive learning techniques~\footnote{Code and visualization at \href{https://contrastive-learning.github.io/intriguing}{https://contrastive-learning.github.io/intriguing}.}.
\end{abstract} \section{Introduction}

Contrastive learning~\cite{becker1992self,dosovitskiy2014discriminative,oord2018representation,wu2018unsupervised,hjelm2018learning,bachman2019learning,henaff2019data,tian2019contrastive,misra2019self,he2019momentum,li2020prototypical,tian2020makes,chen2020simple,chen2020big} has achieved great successes recently for learning visual representations without supervision. 
As shown in~\cite{chen2020simple,chen2020big}, contrastive learning can learn representations that rival supervised learning, and significantly improve the state-of-the-art in semi-supervised learning on ImageNet.
One successful use case of contrastive loss for self-supervised learning is to make \textit{augmented} views of the same example agree~\cite{becker1992self,dosovitskiy2014discriminative,chen2020simple}. A widely used contrastive loss to encourage agreement is based on cross entropy~\cite{sohn2016improved,oord2018representation,wu2018unsupervised,chen2020simple}. Given an augmented view of an example, the contrastive prediction task aims to classify a set of candidates into the positive example (i.e. the other augmented view of the same example) and negative ones via the cross entropy loss. 

In this work, to understand the effectiveness and limitation of existing contrastive learning methods, we study three intriguing aspects. First, we propose a generalization of the standard contrastive loss, and systematically study their performance differences.
Second, we study if the instance-based contrastive learning, for which the contrastive loss operates on global representation of an input image, can learn well on images with multiple objects present, and whether or not it leads to meaningful local features.
Finally, we systematically study the feature suppression phenomenon in contrastive learning. The suppression effect occurs among competing features shared across augmented views. For example, with random cropping as the augmentation, ``color distribution'' and ``object class'' are often competing features as they are likely shared between two augmented views. The suppression effect among competing features can significantly degenerate the representation quality, or even completely disable the learning of certain features, as shown in our experiments. Existing methods critically rely on hand-crafted data augmentation to favor certain sets of competing features than others.

Our main findings and contributions are summarized below.
\begin{itemize}[topsep=0pt, partopsep=0pt, leftmargin=15pt, parsep=0pt, itemsep=5pt]
\item We propose a generalized contrastive loss, and show that differences between contrastive losses are small with a deep projection head.
    \item We show that the instance-based objective widely used in existing contrastive learning methods can learn on images with multiple objects, and also learn meaningful local features despite operating on global image representation.
    \item We construct three datasets with explicit and controllable competing features to systematically study the feature suppression effect in contrastive learning.
    \item We show that a few bits of easy-to-learn shared features can suppress, and even fully prevent, the learning of other sets of competing features. In scenarios where there are multiple objects in an image, the dominant object would suppress the learning of smaller objects. This poses open challenges to existing contrastive learning.
\end{itemize}
 \section{Generalized contrastive loss and differences among its instantiations}
\label{sec:gcl}

The common contrastive loss used in most recent work is based on cross entropy~\cite{sohn2016improved,oord2018representation,wu2018unsupervised}. Following the notation in~\cite{chen2020simple}, the contrastive loss can be defined between two augmented views $(i, j)$ of the same example for a mini-batch of size of $n$, and can be written as the following.

\begin{equation}
\small
\label{eq:nt_xent}
    \mathcal{L}^{\mathrm{NT}\text{-}\mathrm{Xent}} = -\frac{1}{n}\sum_{i,j\in \mathcal{MB}}\log \frac{\exp(\mathrm{sim}(\bm z_i, \bm z_j)/\tau)}{\sum_{k=1}^{2n} \one{k \neq i}\exp(\mathrm{sim}(\bm z_i, \bm z_k)/\tau)}
\end{equation}

where $\bm z_i, \bm z_j$ are hidden representations of two augmented views of the same example; $\mathrm{sim}(\bm u, \bm v)=\bm u^T \bm v/(\|\bm u\|\|\bm v\|)$ is the cosine similarity between two vectors; $\tau$ is a temperature scalar and $\mathcal{MB}$ is a randomly sampled mini-batch consisting of augmented pairs of images. 
In~\cite{chen2020simple}, a MLP \textit{projection head} is introduced between intermediate layer $\bm h$ (e.g. output of ResNet encoder) and final output $\bm z$. It is shown that the projection head is very beneficial and $\bm h$ is a much better feature representation than $\bm z$.

In this work, we generalize the standard contrastive loss to the following form.

\begin{equation}
\label{eq:gc}\boxed{
\mathcal{L}_{\text{generalized contrastive}} = \mathcal{L}_{\text{alignment}} + \lambda \mathcal{L}_{\text{distribution}
}}\end{equation}
~

Both terms are defined on hidden representations. $\mathcal{L}_{\text{alignment}}$ encourages representations of augmented views to be consistent, while $\mathcal{L}_{\text{distribution}}$ encourages representations (or a random subset of them) to match a prior distribution (of high entropy). It is not difficult to see that the standard contrastive loss in Eq.~\ref{eq:nt_xent} is a special case as it can be re-written as follows (scaled by a constant $\tau$).

\begin{equation}
\small
\label{eq:nt_xent_scaled}
\begin{aligned}
    \tau\mathcal{L}^{\mathrm{NT}\text{-}\mathrm{Xent}} = \underbrace{-\frac{1}{n}\sum_{i,j}\mathrm{sim}(\bm z_i, \bm z_j)}_\text{\normalsize $\mathcal{L}_{\text{alignment}}$}
    +  \underbrace{\frac{\tau}{n}\sum_i\log\sum_{k=1}^{2n} \one{k \neq i}\exp(\mathrm{sim}(\bm z_i, \bm z_k)/\tau)}_\text{\normalsize $\mathcal{L}_{\text{distribution}}$}
\end{aligned}
\end{equation}

This form of factorization in Eq.~\ref{eq:nt_xent_scaled} has been proposed in~\cite{wang2020understanding}, 
where the second $\mathrm{LogSumExp}$ term is referred to as \textit{uniformity} since it encourages representation to uniformly distributed in the hypersphere.
Different from~\cite{wang2020understanding}, here we generalize the hypersphere uniform distribution and study a wider set of prior distributions for their effectiveness in learning representations.

\paragraph{SWD for supporting diverse prior distributions.} One issue of using more diverse set of priors is we \textit{cannot} rely on $\mathrm{LogSumExp}$ for matching the distribution. To this end, we resort to the theory of optimal transport, via Sliced Wasserstein Distance (SWD)~\cite{rabin2011wasserstein,bonneel2015sliced,kolouri2019generalized}. For two sets of equal-sized samples from two 1-D distributions, the optimal transport can be obtained by computing two permutations that order the values of both sets of samples respectively. The 1-D Wasserstein distance can then be computed with $\ell_2$ distance between the ordered values. For n-D distributions, we first project the samples to $n$ randomly-generated orthogonal 1-D subspaces, and then compute the sum of 1-D Wasserstein distance across all 1-D subspaces. By adjusting the network weights to minimize the SWD, we are able to reduce the mismatch between the distribution of hidden vectors and a known prior distribution. The detailed algorithm can be found in Algorithm~\ref{alg:swd}.
With SWD loss, we are able to use a wider set of priors, and
Table~\ref{tab:gloss_comp} summarizes instantiations of the generalized contrastive loss with different prior distributions and distribution matching loss.
\begin{algorithm}[!t]
\caption{\label{alg:swd} Sliced Wasserstein Distance (SWD) loss.}
\begin{algorithmic}
    \STATE \textbf{input:} activation vectors $\bm H\in\mathbb{R}^{b\times d}$, a prior distribution (e.g. Gaussian) sampler $\mathcal{S}$
\STATE draw prior vectors $\bm P\in\mathbb{R}^{b\times d}$ using $\mathcal{S}$
    \STATE generate random orthogonal matrix $\bm W\in \mathbb{R}^{d\times d'}$
    \STATE make projections: $\bm H^{\perp} = \bm H\bm W; \bm P^{\perp} = \bm P\bm W$
    \STATE initialize SWD loss $\ell = 0$
    \FOR{$j \in \{1, 2, \cdots, d'\}$}
\STATE $\ell = \ell + \|\mathrm{sort}(\bm H^{\perp}_{:,j})-\mathrm{sort}(\bm P^{\perp}_{:,j})\|^2$
    \ENDFOR
    \STATE \textbf{return} $\ell / (dd')$
\end{algorithmic}
\end{algorithm}

\begin{table*}[!t]
\centering
\small
\caption{\label{tab:gloss_comp} Instantiations of the generalized contrastive loss, i.e. $\mathcal{L}_{\text{alignment}} + \lambda \mathcal{L}_{\text{distribution}}$, that we use in this work. $\tilde{\bm z}$ denotes $\ell_2$-normalized $\bm z\in \mathbb{R}^d$, and is only used for uniform hypersphere prior.}
\begin{tabular}{ccc}
  \toprule
  \textbf{$\mathcal{L}_\text{align}$} & \textbf{Prior distribution} & \textbf{$\mathcal{L}_\text{distribution}$}  \\ \midrule
  $\frac{1}{nd} \sum_{i,j} \|\tilde{\bm z}_i-\tilde{\bm z}_j\|^2$ & Uniform hypersphere & $\frac{1}{n}\sum_i \log\sum_{j} \exp(\tilde{\bm z_i}^T \tilde{\bm z_j}/\tau)$     \\ \midrule $\frac{1}{nd} \sum_{i,j}\|\tilde{\bm z}_i-\tilde{\bm z}_j\|^2$  & Uniform hypersphere & SWD($\tilde{Z}, Z^{prior}$)     \\ \midrule 
  $\frac{1}{nd} \sum_{i,j}\|\bm z_i-\bm z_j\|^2$  & Uniform hypercube & SWD($Z, Z^{prior}$)     \\ \midrule 
  $\frac{1}{nd} \sum_{i,j}\|\bm z_i-\bm z_j\|^2$  & Normal distribution & SWD($Z, Z^{prior}$)     \\
  \bottomrule
\end{tabular}
\end{table*}

\paragraph{Connection with mutual information.} The connection between the standard contrastive loss and mutual information has been shown before~\cite{oord2018representation,poole2019variational}, where the contrastive loss (a.k.a. InfoNCE loss~\cite{oord2018representation}) is shown to be a lower bound of the mutual information. To connect the generalized contrastive loss to mutual information, we start by the definition of mutual information between two latent variables $U, V$, which is $I(U; V) = H(U) - H(U|V)$.
Comparing this factorization of mutual information with generalized contrastive loss, it is not difficult to see that: 1) the alignment term $\mathcal{L}_{\text{alignment}}$ is directly related to $H(U|V)$ which aims to reduce uncertainty of the other views given one view of the example; and 2) the distribution matching term $\mathcal{L}_{\text{distribution}}$ can be considered as a proxy to $H(u)$ for maximizing the entropy in the representation.
It is perhaps worth noting that different from mutual information, the generalized contrastive loss (Eq.~\ref{eq:gc}) allows a tunable weight ($\lambda$) between the alignment and distribution matching term. The weighting scalar $\lambda$ is (inversely) related to the temperature $\tau$ (details in Appendix~\ref{app:tau_lambda}).

\begin{figure*}[t]
    \centering
    \begin{subfigure}{.32\textwidth}
      \centering
      \includegraphics[width=0.9\linewidth]{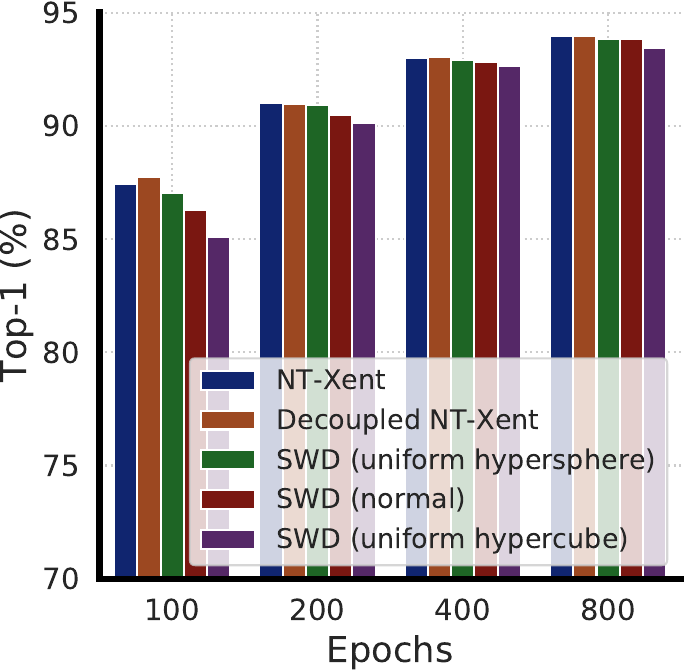}
      \caption{CIFAR-10 (2 layers)}
    \end{subfigure}
    \begin{subfigure}{.32\textwidth}
      \centering
      \includegraphics[width=0.9\linewidth]{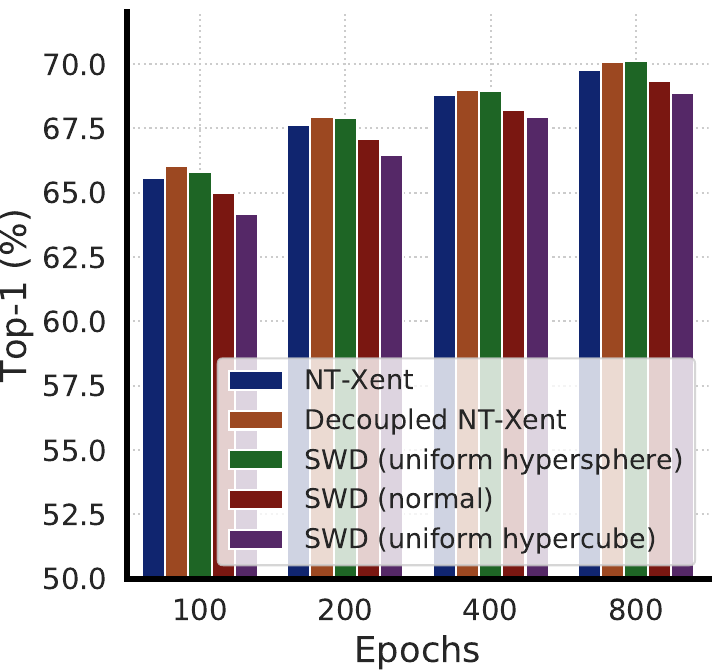}
      \caption{ImageNet (2 layers)}
    \end{subfigure}
    \begin{subfigure}{.32\textwidth}
      \centering
      \includegraphics[width=0.9\linewidth]{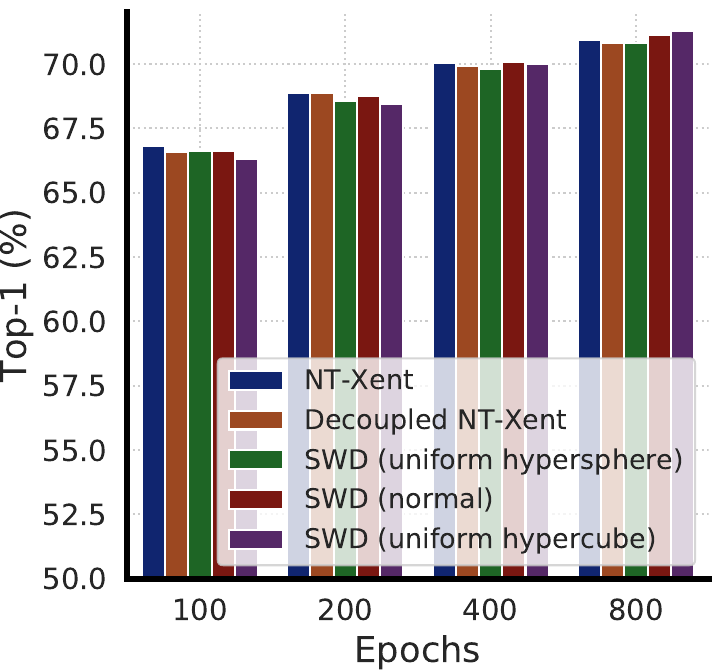}
      \caption{ImageNet (3 layers)}
    \end{subfigure}
    \caption{\label{fig:comploss_linear}Linear evaluation accuracy of ResNet-50 trained with different losses on CIFAR-10 and ImageNet datasets.
    Numbers of projection head layers are in parentheses. Differences between variants of generalized contrastive loss are small with a deep projection head. Decoupled NT-Xent loss is introduced in~\ref{app:tau_lambda}. Numerical results can be found in Appendix~\ref{app:linear_eval}.}
\end{figure*}

\paragraph{Comparing different instantiations of generalized contrastive loss.} Here we ask: Is it essential to use a uniform hypersphere prior for the effectiveness of contrastive loss? How much differences does it make when different generalized contrastive losses are used? To answer this question, we conduct experiments following SimCLR settings~\cite{chen2020simple,chen2020big}, and use the linear evaluation protocol. Detailed experimental setup can be found in appendix~\ref{app:gloss_setup}.

Figure~\ref{fig:comploss_linear} shows linear evaluation results of models trained with different losses under different training epochs. On CIFAR-10, we see little differences in terms of linear evaluation for variants of the generalized contrastive losses, especially when trained longer than 200 epochs. As for ImageNet, there are some discrepancies between different losses, but they disappear when a deeper 3-layer non-linear projection head is used.

\begin{wraptable}{r}{0.62\textwidth}
    \centering
    \small
\caption{\label{tab:linear_eval_num_imagenet}Linear eval accuracy of ResNet-50 on ImageNet.}
    \begin{tabular}{cc|rrrr}
    \toprule
          \multirow{2}{*}{Projection head}          &       \multirow{2}{*}{Batch size}   & \multicolumn{4}{c}{Epoch}  \\
     &   &     100 &   200 &   400 &   800 \\
    \midrule
    \multirow{3}{*}{2 layers} & 512  &  65.4 &  67.3 &  68.7 &  69.3 \\
                              & 1024 &  65.6 &  67.6 &  68.8 &  69.8 \\
                              & 2048 &  65.3 &  67.6 &  69.0 &  70.1 \\
    \midrule
    \multirow{3}{*}{3 layers} & 512  &  66.6 &  68.4 &  70.0 &  71.0 \\
                              & 1024 &  66.8 &  68.9 &  70.1 &  70.9 \\
                              & 2048 &  66.8 &  69.1 &  70.4 &  71.3 \\
    \midrule
    \multirow{3}{*}{4 layers} & 512  & 66.8 & 68.8 & 70.0 & 70.7 \\
                              & 1024 & 67.0 & 69.0 & 70.4 & 70.9 \\
                              & 2048 & 67.0 & 69.3 & 70.4 & 71.3 \\
    \bottomrule
    \end{tabular}
    \vspace{-1.0em}
\end{wraptable}

Furthermore, we find that deep projection head not only reduces the differences among different generalized contrastive losses, but has a similar effect for batch size. With proper learning rate scaling across batch sizes (e.g. square root scaling with LARS optimizer~\cite{you2017large}), the impact of batch size on representation quality is small. Table~\ref{tab:linear_eval_num_imagenet} demonstrate this phenomenon for the standard contrastive loss, and more results on other losses can be found in Appendix~\ref{app:linear_eval}. \section{Instance-based objective can learn on images with multiple objects and learn good local features}

Most existing contrastive learning methods~\cite{chen2020simple,he2019momentum,grill2020bootstrap,wu2018unsupervised} define their objectives at the instance level where each image is encoded into a single vector representation (e.g. representations of two random crops of the same image instance are treated as a positive pair). In other words, the objective operates on a global representation of its input rather than on some local regions (of its input).
We pose two questions regarding instance-based global objective: 1) when there is only a single (dominant) object in the image, the objective seems reasonable as it encourages the model to learn features relevant to object class, but when there are multiple objects present in the image, can instance-based objective still learn well? 2) Since the instance-based objective uses a global summary of its input, can it still learn good local features (e.g. parts of an object, or multiple objects in the same scheme)? To answer these questions, we use SimCLR as representative for the instance-based objective.

\subsection{SimCLR can learn on images with multiple objects}

Commonly used self-supervised learning datasets, such as MNIST, CIFAR-10, ImageNet, are object centered, i.e. the image is mainly occupied by a single (dominant) object. To experiment with multiple objects in a controllable setting, we propose a new dataset setting by composing multiple digits as follows.

\begin{figure*}[!t]
    \centering
    \begin{subfigure}{.495\textwidth}
      \centering
      \includegraphics[width=0.95\linewidth]{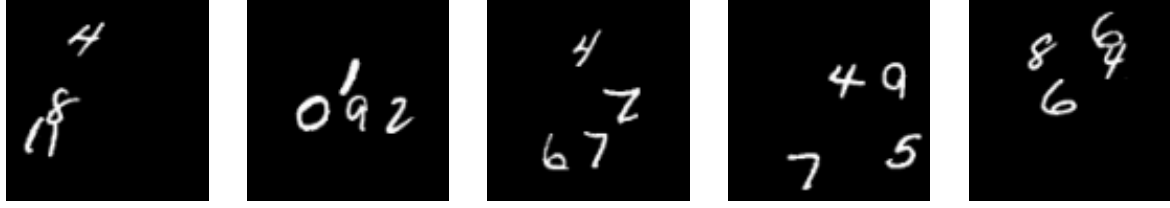}
      \caption{4 digits, random placement.}
    \end{subfigure}
    \begin{subfigure}{.495\textwidth}
      \centering
      \includegraphics[width=0.95\linewidth]{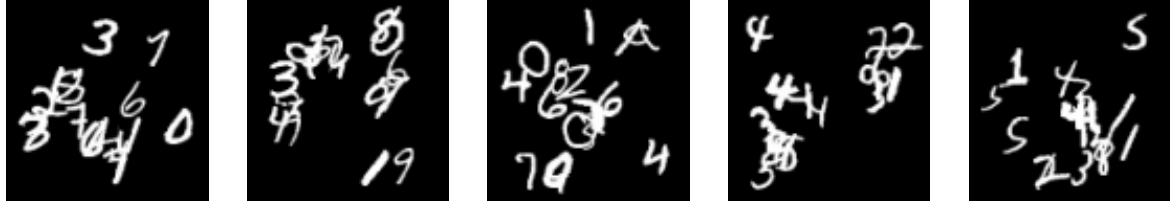}
      \caption{16 digits, random placement.}
    \end{subfigure}
    \begin{subfigure}{.495\textwidth}
      \centering
      \includegraphics[width=0.95\linewidth]{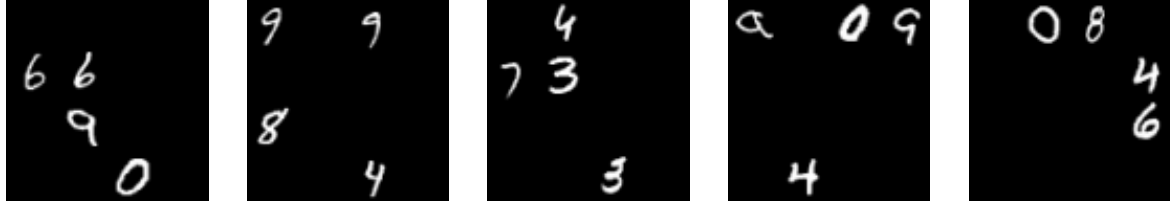}
      \caption{4 digits, in-grid placement.}
    \end{subfigure}
    \begin{subfigure}{.495\textwidth}
      \centering
      \includegraphics[width=0.95\linewidth]{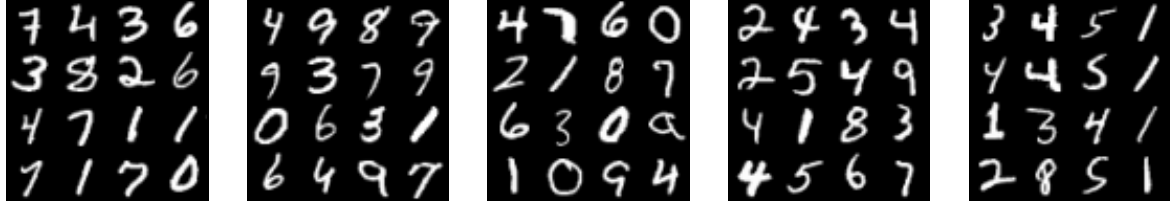}
      \caption{16 digits, in-grid placement.}
    \end{subfigure}
    \caption{\label{fig:probe_data_multidigit}MultiDigit dataset. More digits lead to more overlapping in random placement.}
\end{figure*}

\textbf{MultiDigits dataset}. 
We place MNIST digits ($28\times28$ size) on a shared canvas ($112\times112$ size). We vary the number of digits placed on the canvas. One factor that could interfere with learning of multiple digits is overlapping digits, therefore we use two placement strategies: random vs in-grid (Figure \ref{fig:probe_data_multidigit}). Random placement of digits incurs no constraint on where digits can be placed on the canvas, whereas in-grid placement puts each digit in one of the $4\times4$ grid cells the canvas is divided into, and no two digits can fall in the same cell. In-grid placement ensures no overlapping of digits.

We first pretrain a ResNet-18 with SimCLR or supervised learning with the same augmentation policy (random cropping and resize) 
on MultiDigits dataset.
To access the representation quality, we then train linear classifiers for images with a single digit of size $28\times28$ on the canvas. Similarly during evaluation, we place only one digit of size $28\times28$ on the canvas.

As shown in Table \ref{tab:multidigit}, representations learned using supervised loss maintains its quality when up to 8 digits are placed in the image. After that the representation becomes worse as the canvas gets more crowded. Notably, representations learned using SimCLR display a similar phenomenon. Regardless of placement strategy, top-1 accuracy stays at the same level up to 8 digits, demonstrating that SimCLR can learn from images with multiple objects. In addition, the increased performance gap between the two placement strategies with increased number of digits shows that object overlapping makes it harder for contrastive losses to learn from multiple objects.

\begin{table*}[h]
\centering
\small
\caption{\label{tab:multidigit} Top-1 linear evaluation accuracy (\%) for pretrained ResNet-18 on the MultiDigits dataset.
We vary the number of digits placed on the canvas during training from 1 to 16. During evaluation only 1 digit is present.
As a baseline, a network with random weights gives 18\% top-1 accuracy.}
\begin{tabular}{llcccccc}
  \toprule
   & Placing of digits & \multicolumn{6}{c}{Number of digits (size $28\times28$)} \\
   & & 1 & 2 & 4 & 8 & 12 & 16 \\
  \midrule
  \multirow{2}{*}{Supervised} & Random & 99.5 & 99.5 & 99.3 & 99.4 & 98.9 & 98.3 \\
             & In-grid & 99.5 & 99.6 & 99.5 & 99.3 & 98.6 & 92.4 \\
  \midrule
  \multirow{2}{*}{SimCLR} & Random & 98.9 & 98.9 & 99.0 & 98.9 & 98.2 & 96.4 \\
             & In-grid & 98.3 & 98.6 & 99.1 & 99.2 & 99.1 & 98.3 \\
  \bottomrule
\end{tabular}
\end{table*}

\subsection{SimCLR learns local features that exhibit hierarchical properties}

To understand the local features learned by SimCLR, we apply K-means on intermediate features of the pretrained ResNet with SimCLR, and see how local regions of an image are grouped together. For good representations, we expect that regions of similar objects or object parts should be grouped together.

Specifically, we take a pretrained Resnet-50 $2\times$ 
on ImageNet, and run inference on images (from ImageNet validation set and COCO~\cite{lin2014microsoft}) of size 448$\times$448. We run K-means with various numbers of clusters on the l2-normalized hidden features from middle layers of the network (e.g. block group 2,3,4 of the ResNet). We also compare SimCLR learned features with supervised learned features, as well as the raw pixel (RGB) features extracted from each $14\times14$ patch.

\begin{figure*}[!t]
    \centering
    \begin{subfigure}{.495\textwidth}
      \centering
      \includegraphics[width=0.99\linewidth]{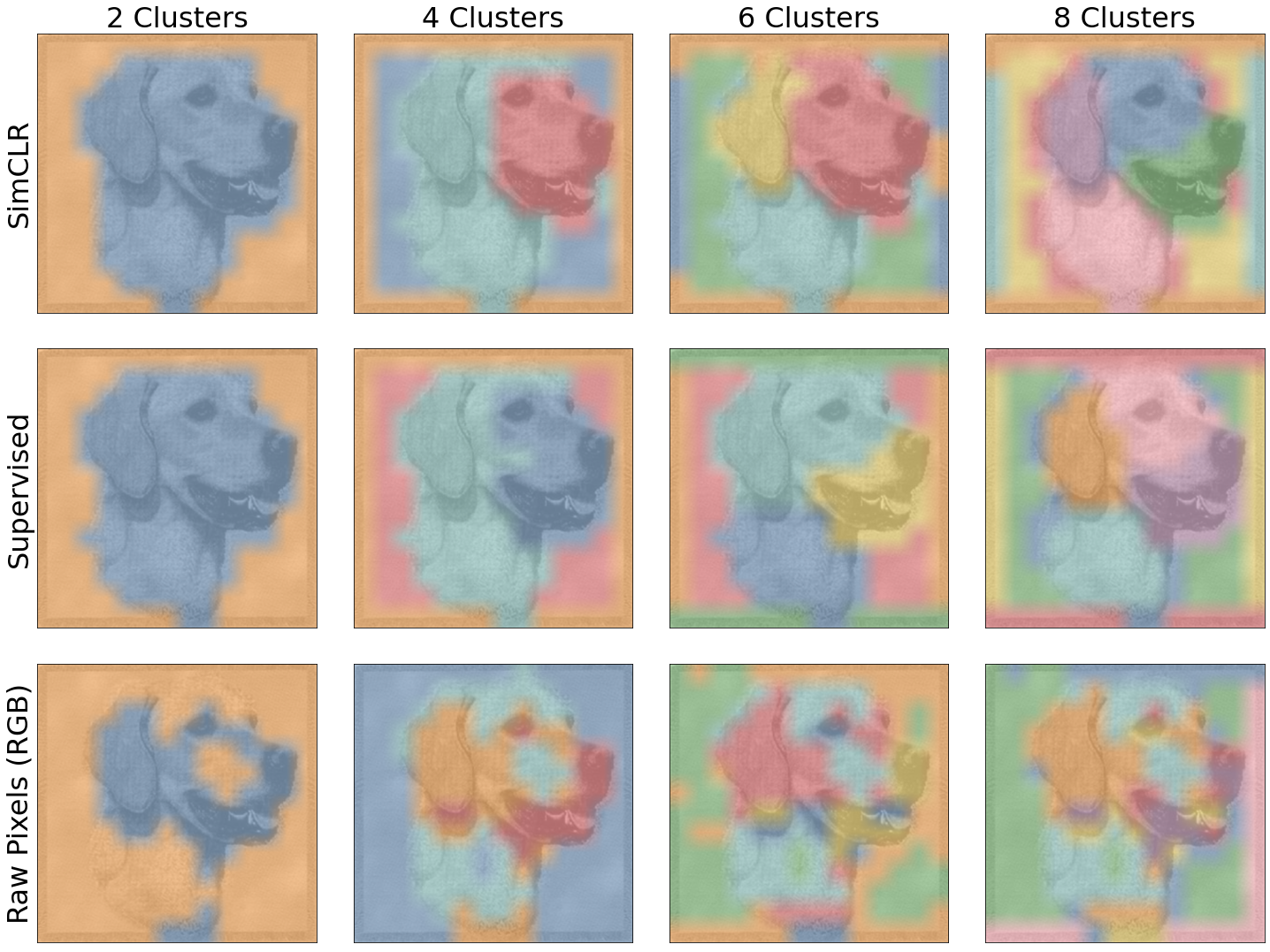}
      \caption{\label{fig:vis_local_parts_a}Features from different methods.}
    \end{subfigure}
    \begin{subfigure}{.495\textwidth}
      \centering
      \includegraphics[width=0.99\linewidth]{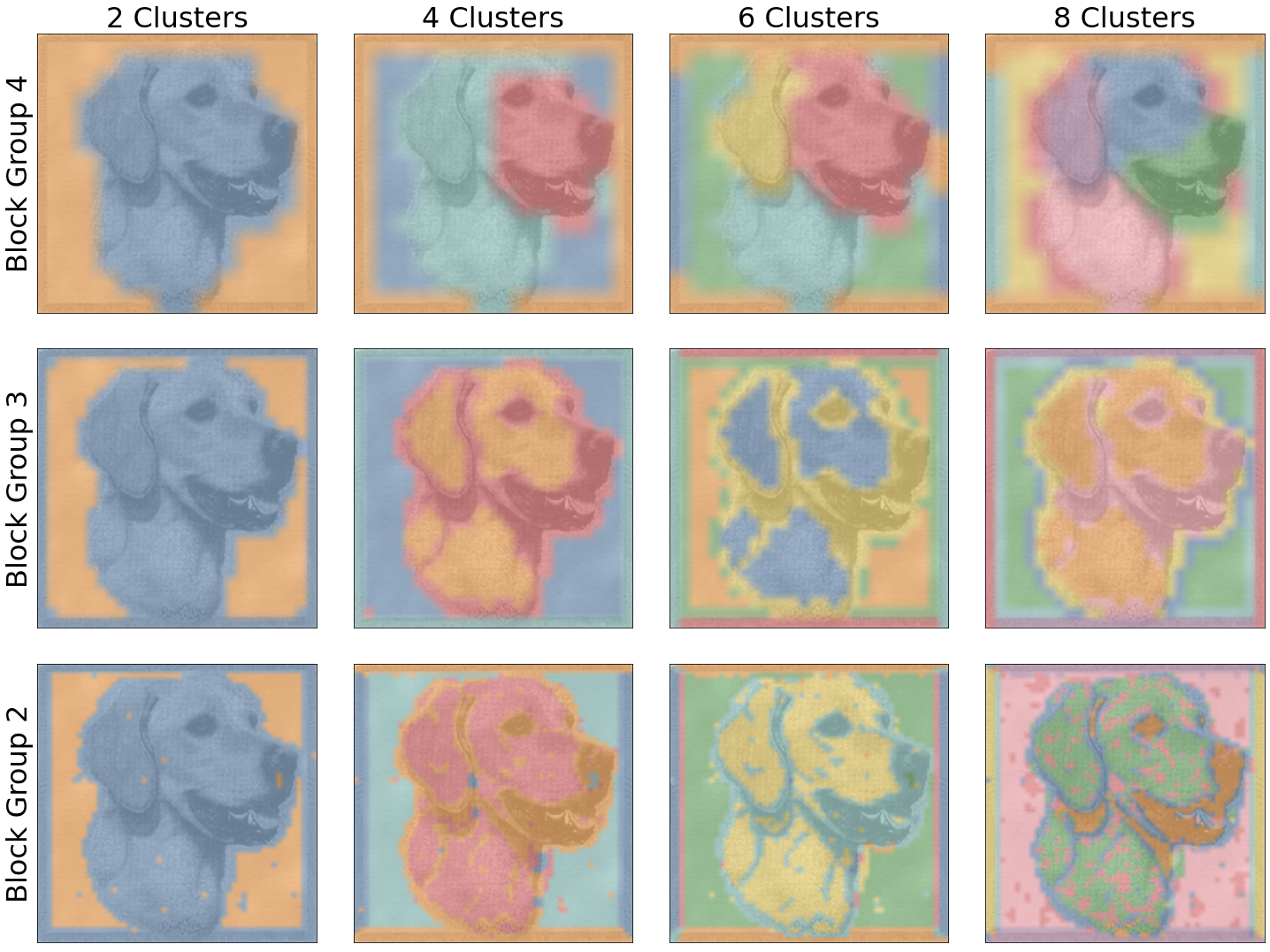}
      \caption{\label{fig:vis_local_parts_b}SimCLR features at different ResNet layers.}
    \end{subfigure}
    \caption{\label{fig:vis_local_parts}Visualizing features on a ImageNet validation image with K-means clustering. Each row denotes a type of local features used, and each column denotes the number of K-means clusters. 
    Later layers of SimCLR/supervised ResNet tend to group by object parts.
    More visualization examples can be found in \href{https://contrastive-learning.github.io/intriguing}{https://contrastive-learning.github.io/intriguing}.
    }
\end{figure*}

Figure~\ref{fig:vis_local_parts_a} shows that as the number of clusters increases, the learned representations tend to group image regions based on parts of the object (i.e. facial components of the dog). This phenomenon appears in both SimCLR and supervised learned features, but not with raw pixel features, indicating meaningful local features learned by SimCLR and supervised learning. In Figure~\ref{fig:vis_local_parts_b}, we compare ResNet intermediate features at different layers, and it suggests that earlier layers contain more edge-related features, while later layers contain more object/part features.

\begin{figure*}[!t]
    \centering
    \begin{subfigure}{.495\textwidth}
      \centering
      \includegraphics[width=0.99\linewidth]{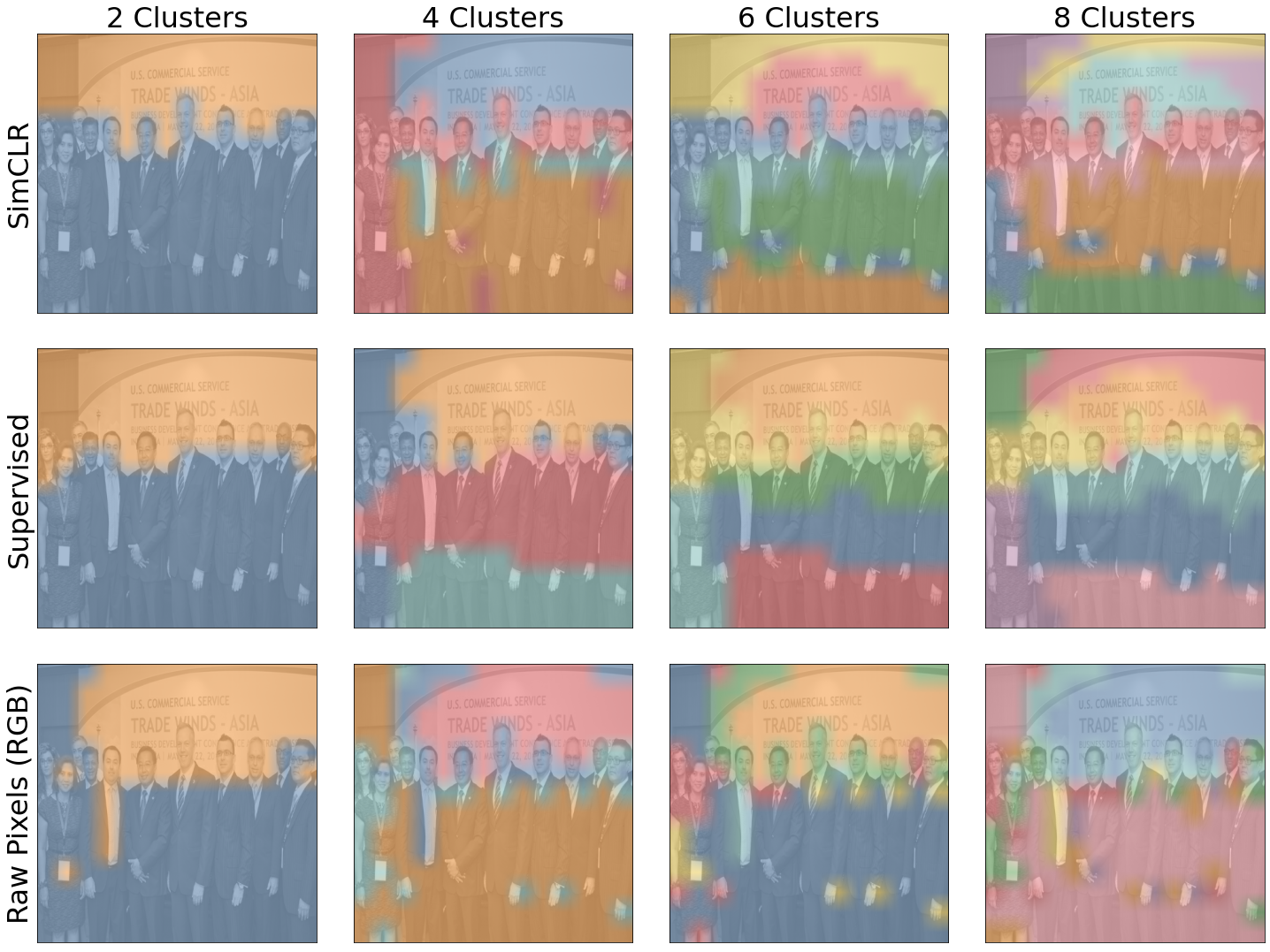}
      \caption{}
    \end{subfigure}
    \begin{subfigure}{.495\textwidth}
      \centering
      \includegraphics[width=0.99\linewidth]{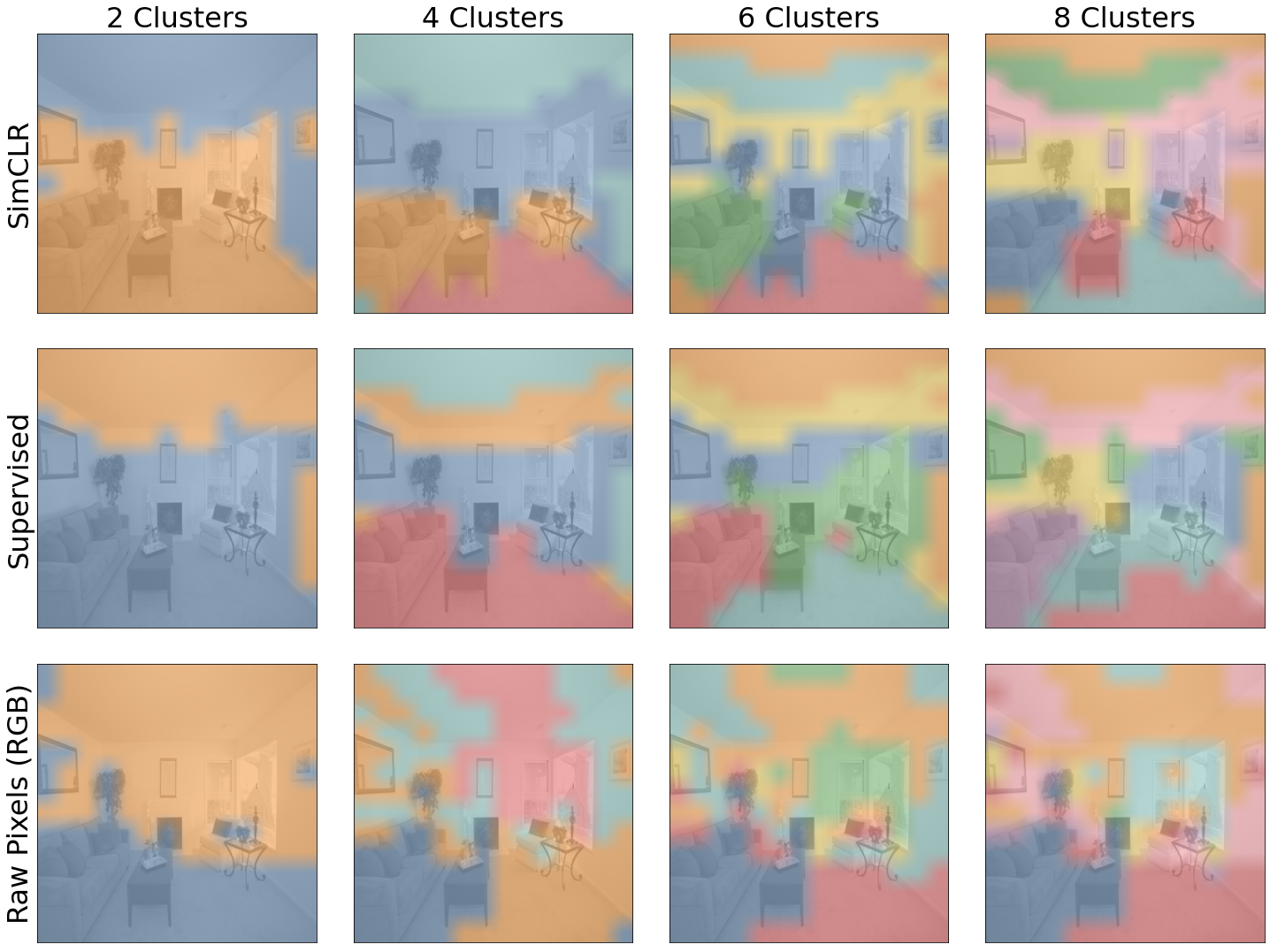}
      \caption{}
    \end{subfigure}
    \caption{\label{fig:vis_local_multi_objs}Visualizing features on two images from COCO. Each row denotes a type of local features (SimCLR, Supervised, and raw pixels; both SimCLR and Supervised are trained on ImageNet), and each column denotes the number of K-means clusters. Region grouping by SimCLR/supervised features tend to overlap with object class.}
\end{figure*}

Figure~\ref{fig:vis_local_multi_objs} show region grouping results on two COCO images of SimCLR and supervised learning (trained on ImageNet). Again, region grouping by local features tend to overlap with object class, indicating good local features learned.
 \section{Feature suppression limits the potential of contrastive learning}
\label{sec:cf}

\begin{figure*}[!t]
    \centering
    \begin{subfigure}{.9\textwidth}
      \centering
      \includegraphics[trim={1cm 0 0 0},clip,width=0.82\linewidth]{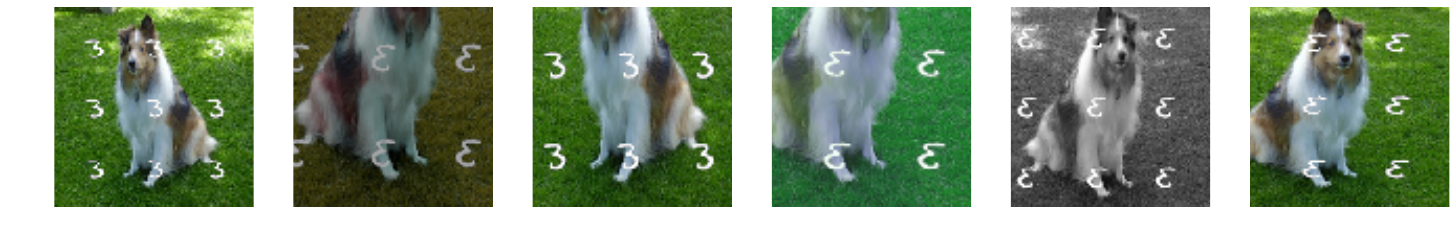}
      \includegraphics[trim={1cm 0 0 0},clip,width=0.82\linewidth]{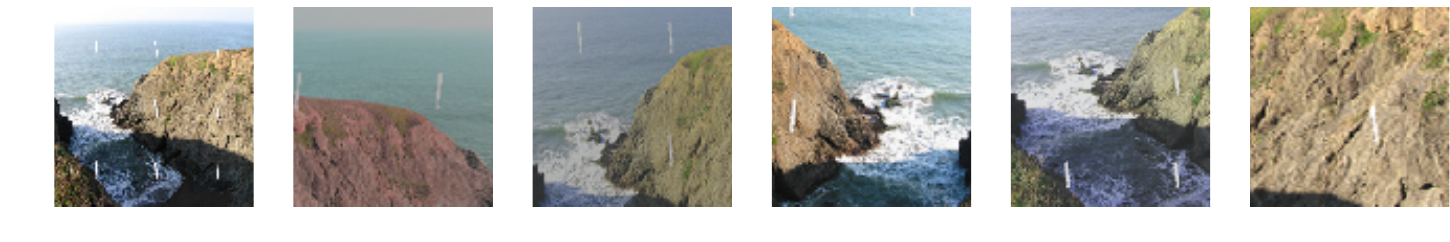}
      \caption{\label{fig:probe_data_digits}ImageNet images overlaid with MNIST digits. The left most column is original image, and others are augmented views via random crop and color distortion. MNIST digits and ImageNet classes are competing features. We vary the number of unique MNIST digits to control the competing features.}
    \end{subfigure}
    ~~\\~~\\~~\\
    \begin{subfigure}{.9\textwidth}
        \centering
        \includegraphics[width=0.83\textwidth]{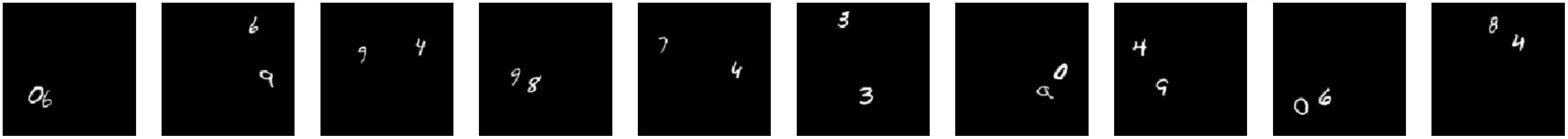}
        \includegraphics[width=0.83\textwidth]{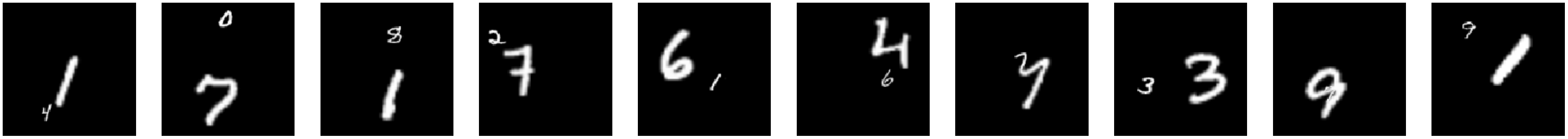}
        \caption{\label{fig:probe_data_2digit}Two MNIST digits randomly placed on a shared canvas (of size $112 \times 112$). 
The two digits can have the same size (upper row) or different sizes (lower row), and digits of different sizes can be considered as competing features. We fix the size of one digit and vary the other.}
    \end{subfigure}
    ~~\\~~\\~~\\
    \begin{subfigure}{.9\textwidth}
        \centering
        \includegraphics[width=0.75\textwidth]{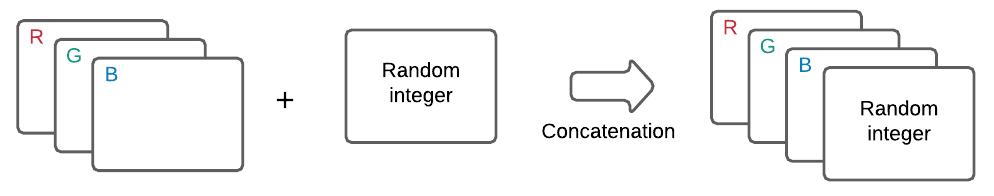}
        \caption{\label{fig:probe_data_rgba}Images (of RGB channels) are concatenated with additional channels of random integer sampled from range of $[1, \log_2(n)]$. The integer, shared between two views, is replicated for spatial dimension and represented as $n$ binary channels. RGB channels and random bits are competing features.}
    \end{subfigure}
    \caption{\label{fig:probe_data}Probing datasets with explicit and controllable competing features.}
\end{figure*}

Contrastive learning requires good design of data augmentation to work well. 
As shown in~\cite{chen2020simple}, without color augmentation that randomly shift color distribution (while maintaining information regarding object class), the quality of learned representations are significantly worse. 
In other words, the presence of ``color distribution'' features suppresses their competing feature of ``object class'', and is addressed by color augmentation.
However, there may be scenarios where the known augmentations cannot fully address this feature suppression effect, and it can thus limit the potential of contrastive learning.
Here we quantitatively study the feature suppression phenomenon by constructing datasets with explicit and controllable competing features, and see how well contrastive learning method could learn.

\subsection{Datasets with explicit and controllable competing features}

To construct datasets with controllable competing features, we leverage two strategies: \textit{channel addition} that adds different feature information in a shared canvas, and \textit{channel concatenation} that expand the RGB channels to include additional features. With these strategies, we construct three datasets below.

\textbf{DigitOnImageNet dataset}. We overlay MNIST digits on ImageNet images via channel addition/summation (Figure~\ref{fig:probe_data_digits}).
For each ImageNet image, we assign an unique MNIST digit and replicate it in nine fixed locations before the standard SimCLR augmentations~\cite{chen2020simple} are applied to create augmented views. Therefore the original ImageNet images and added MNIST digits are competing features. Although it is difficult to quantify information in MNIST digits, we can manually control the number of unique MNIST digits used. Ideally, we want the model to learn both set of features so that it could perform well for both MNIST digit and ImageNet object recognition.

\textbf{MultiDigits dataset (varying the size of one digit)}. 
This dataset is modified from MultiDigits introduced above. Here we only consider two digits and varying the size of one digit (Figure~\ref{fig:probe_data_2digit}). 
The canvas size is $112\times112$, and we place two digits on it in this work. We fix the size of one of the digits to be $20\times 20$ while varying the other from $20\times20$ to $80\times80$. Digits of different sizes can be considered as competing features. Ideally, we want the model to learn features for digits of all sizes appeared during training.

\textbf{RandBit dataset}. We concatenate a real image with an image of a random integer in their channel dimension (Figure~\ref{fig:probe_data_rgba}). The random integer is randomly sampled from range of $[1, \log_2(n)]$ where $n$ is a parameter to control. It is replicated across spatial dimension (i.e. all pixel location shares the same value), and it is also represented as $n$ binary bits/channels instead of an integer or floating number to make it easily learnable. Furthermore, unlike RGB channels, these additional channels of random bits will \textit{not} be altered by augmentation, so they are identical for both augmented views of the same image.
The RGB channels and the added channels of random bits are competing features, and this construction allows us to control the amount of information in the added competing feature, which is $n$ bits. Also, we know that the mutual information between two views given this construction is at least $\log_2(n)$.

\subsection{Easy-to-learn features (MNIST digit) suppress the learning of other features (ImageNet object class)}

\begin{figure*}[h]
    \centering
    \begin{subfigure}{.25\textwidth}
      \centering
      \includegraphics[width=0.98\linewidth]{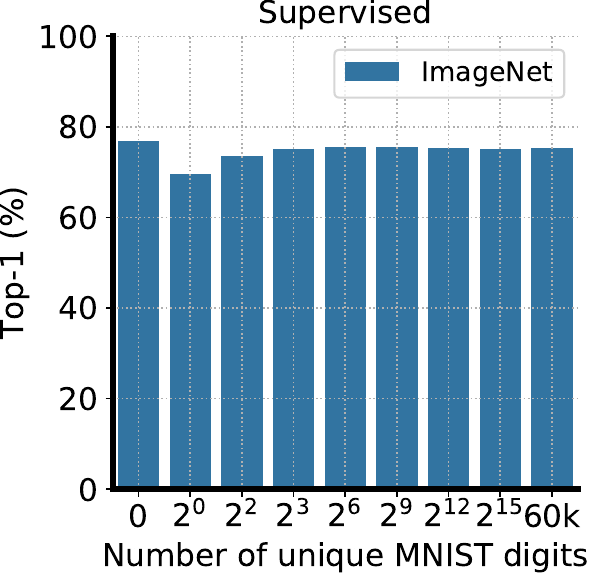}
      \caption{\label{fig:linear_overlay_sup}Supervised learning}
    \end{subfigure}
    \begin{subfigure}{.695\textwidth}
      \centering
      \includegraphics[width=0.98\linewidth]{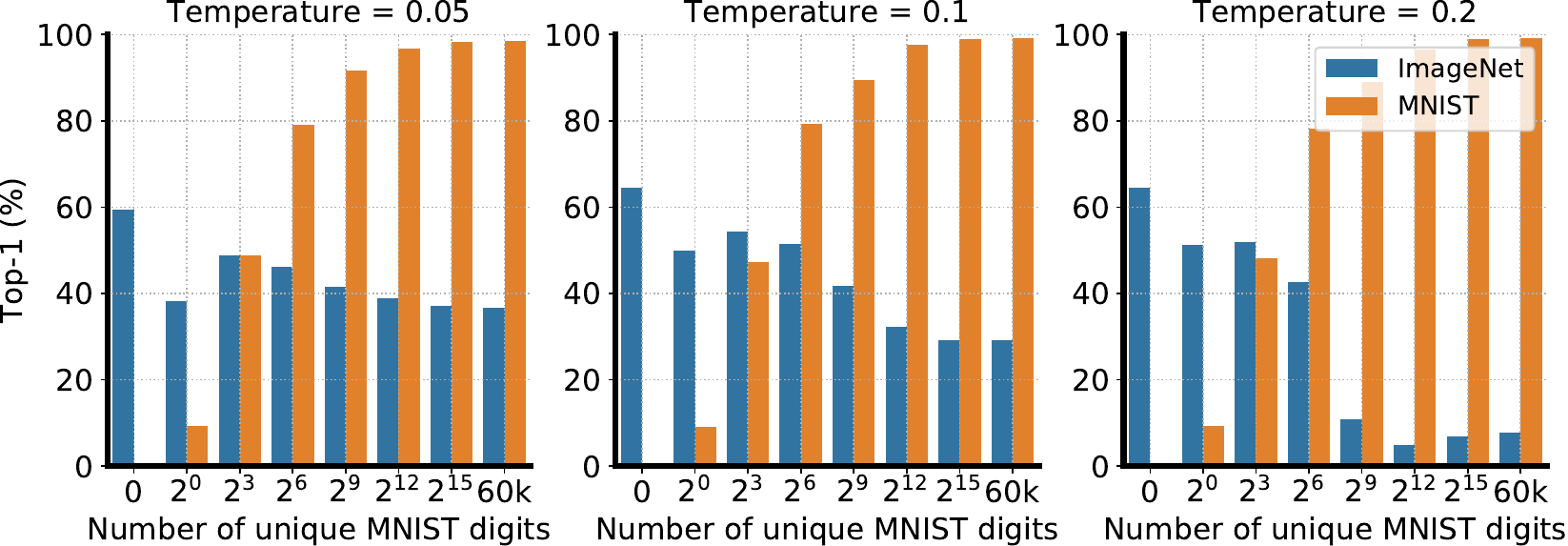}
      \caption{\label{fig:linear_overlay_unsup}Unsupervised contrastive learning}
    \end{subfigure}
    \caption{\label{fig:linear_overlay}
    (a) Supervised learning accuracy on ImageNet classification.
    (b) Linear evaluation of learned features for both MNIST classification and ImageNet classification on the DigitOnImageNet dataset. 
Batch size of 1024 and 2-layer projection head is used. Different batch sizes and projection head layers have negligible influence on the trade-off between ImageNet vs MNIST accuracy.}
\end{figure*}

On DigitOnImageNet datasets, we vary the number of unique MNIST digits used in the training set, and all MNIST digits are used in the validation/test set.
As a baseline, we train supervised ResNet-50 on the created datasets with ImageNet labels, and the number of unique MNIST digits has little impact on the top-1 ImageNet classification accuracy (Figure~\ref{fig:linear_overlay_sup}). 

We then train SimCLR on the datasets with different temperatures. As shown in Figure~\ref{fig:linear_overlay_unsup}, when we increase the number of unique MNIST digits, the linear evaluation performance of the learned features for MNIST classes increases accordingly, while the accuracy for ImageNet classes decreases dramatically. {The trade-off between digit recognition ability and object recognition ability shows that simple features suppress the learning of difficult features, when both are shared between two augmented views}. 
Different batch sizes and projection head depths have negligible influence to the outcome we observe here. 
Therefore, it is difficult to learn both of the competing features using existing contrastive losses (e.g. SimCLR).

\subsection{The presence of dominant object suppresses the learning of features of smaller objects}

On the MultiDigits dataset, as mentioned, we fix one digit to be size of $20\times20$ while varying the other from $20\times20$ to $80\times80$, on a canvas of $112\times112$. We first pretrain a ResNet-18 with SimCLR or supervised learning with the same augmentation policy (random cropping and resize) and batch size of 1024. To access the representation quality, we then train linear classifiers for each of the digit sizes that appeared during pretraining. For training of the linear classifier, we only place a single digit at a time on the canvas of the same size as during pretraining. 

\begin{table*}[h]
\centering
\small
\caption{\label{tab:2digit} Top-1 linear evaluation accuracy (\%) for pretrained ResNet-18 on the MultiDigits dataset. 
We fix the size of 1st digit while increasing the size of the 2nd digit. For SimCLR, results are presented for two temperatures. Accuracies suffered from a significant drop when increasing 2nd digit size are red colored.
}
\begin{tabular}{llccccccc}
  \toprule
   & & \multicolumn{7}{c}{2nd digit size (1st digit is kept the same size of $20\times20$)} \\
   & & $20\times20$ & $30\times30$ & $40\times40$ & $50\times50$ & $60\times60$ & $70\times70$ & $80\times80$ \\
  \midrule
  \multirow{2}{*}{Supervised} & 1st digit & 99.1 & 99.2 & 99.2 & 99.2 & 99.1 & 99.1 & 99.0 \\
             & 2nd digit & 99.1 & 99.5 & 99.5 & 99.6 & 99.5 & 99.5 & 99.6 \\
  \midrule
  SimCLR  & 1st digit & 97.8 & 97.6 & 96.2 & 96.5 & \textcolor{red}{88.5} & \textcolor{red}{74.5} & \textcolor{red}{39.9} \\
  ($\tau=0.05$) & 2nd digit & 97.8 & 97.9 & 97.8 & 98.3 & 98.2 & 97.7 & 98.2 \\
  \midrule
  SimCLR & 1st digit& 98.7 & 98.8 & 98.3 & \textcolor{red}{87.5} & \textcolor{red}{24.9} & \textcolor{red}{19.8} & \textcolor{red}{20.3} \\
  ($\tau=0.2$) & 2nd digit & 98.7 & 99.2 & 99.2 & 99.0 & 99.1 & 98.9 & 99.4 \\
  \midrule
  Random net & 1st digit  & 16.5 & 16.7 & 16.6 & 16.6 & 16.6 & 16.9 & 16.5 \\
  (untrained)        & 2nd digit & 16.5 & 19.1 & 21.9 & 24.1 & 26.5 & 28.1 & 29.0 \\
  \bottomrule
\end{tabular}
\end{table*}

The results are summarized in Table~\ref{tab:2digit}. For supervised learning, the learned representations for the smaller digit do not change much as the other digit increases its size, and the model perform well for both small and large digits (accuracy $>99\%$). However, for SimCLR, the learned representations of the smaller digit degenerate significantly when the size of the other digit increases, almost to the level of a random untrained network. The dominant object can be learned very well (accuracy $>99\%$) while suppressing the learning of the smaller object. Although tuning temperature has some effects on reducing the feature suppression, the trend stays unchanged.

\subsection{Extra channels with a few bits of easy-to-learn mutual information suppress the learning of all features in RGB channels}

In the RandBit datasets, we add additional channels (identical across pixels) of random bits to MNIST and ImageNet. As mentioned above, SimCLR augmentation is only applied to RGB channels so extra added channels will be shared among two view. 

\begin{figure}[h]
    \centering
    \begin{subfigure}{.71\textwidth}
      \centering
      \includegraphics[width=0.98\linewidth]{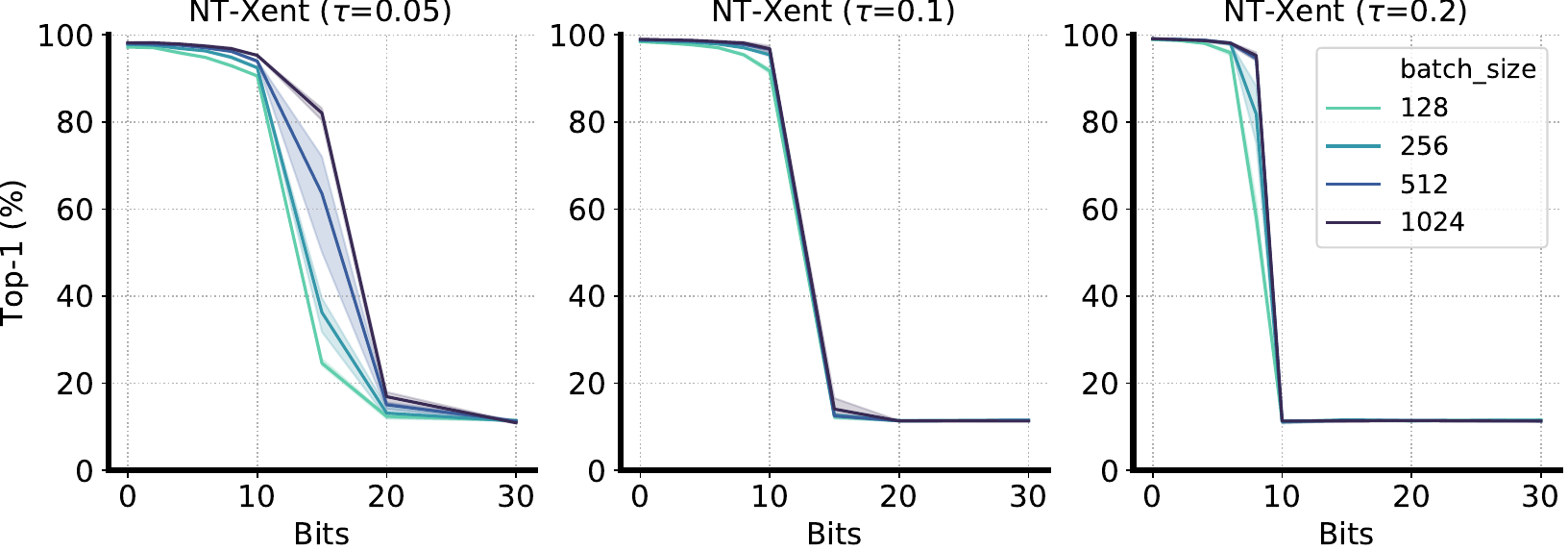}
      \caption{\label{fig:linear_randb_mnist}Contrastive loss.}
    \end{subfigure}
    \begin{subfigure}{.26\textwidth}
      \centering
      \includegraphics[width=0.98\linewidth]{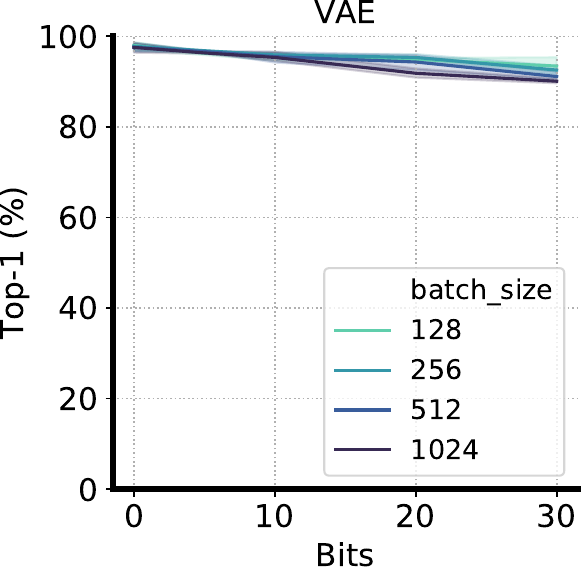}
      \caption{\label{fig:linear_randb_mnist_vae}Generative loss.}
    \end{subfigure}
\caption{\label{fig:linear_randb} Linear evaluation of learned features when a few bits of competing features added (on MNIST). A few bits added completely disable contrastive learning (across various batch size or losses). Interestingly, it has little effects on a generative model (VAE). The detrimental effects are just as strong for larger datasets such as CIFAR-10 and ImageNet (Appendix~\ref{app:ext_probing_result_bit}).
}
\end{figure}
Figure~\ref{fig:linear_randb} shows the linear evaluation accuracy of models trained on MNIST (with additional random bits added). 
We observe that the linear evaluation accuracy quickly drops with a few bits of competing feature added. This detrimental effect on the representation quality persists on bigger datasets like CIFAR-10 and ImageNet as well, and cannot be avoided by using different contrastive losses, batch sizes, or memory mechanism based on momentum contrast (details in Appendix~\ref{app:ext_probing_result_bit}).
We believe the fact that just a few bits of easy-to-learn features can completely disable the good representation learning is related to the saturation of the distribution matching loss. As shown in Appendix~\ref{app:dist_sat}, the linear increase in bits requires an exponential increase in batch size, which is not sustainable as the required batch size can quickly go beyond the size of the dataset size. In practice, we rely on using data augmentation to remove those uninformative easy-to-learn features so that contrastive learning can learn useful representations.
Interestingly, the extra bits do not affect a generative model, variational autoencoder~\cite{kingma2013auto,rezende2014stochastic}, nearly as much, despite other settings such as model size are held the same, prompting a potential direction of addressing the issue.
 \section{Related Work}
Our work studies the contrastive loss based on cross entropy loss~\cite{sohn2016improved,oord2018representation,wu2018unsupervised,chen2020simple}. This loss is widely used in recent successful contrastive learning methods~\citep{oord2018representation,wu2018unsupervised,hjelm2018learning,bachman2019learning,henaff2019data,tian2019contrastive,misra2019self,he2019momentum,li2020prototypical,tian2020makes,chen2020simple,chen2020big}. In terms of the contrastive loss, our work is perhaps most related to~\cite{wang2020understanding}, which shows that formulating contrastive loss as alignment and uniformity in the hypersphere gives similar performance as the standard contrastive loss. We further generalize this factorization, and show other distribution matching losses can be used, and they could achieve similar results. Other than standard contrastive loss that directly utilize negative examples, BYOL ~\cite{grill2020bootstrap} demonstrates another way to maintain representation distribution/entropy without directly relying on distribution matching, and SWAV~\cite{caron2020unsupervised} shows clustering-based method equipped with proper data augmentations could also achieve similar performance. We conduct preliminary experiments of BYOL on RandBit and found that it also suffers from feature suppression as generalized contrastive loss. It is expected that SWAV would exhibit similar behaviors on RandBit as those random bits could fuel representations for perfect clustering. 

The connection between contrastive loss and mutual information has been studied before~\cite{oord2018representation,poole2019variational}. We show that for the generalized contrastive loss, it can also be related to mutual information. Despite the connection between contrastive loss and mutual information, it has been pointed out that mutual information estimation may suffer from certain limitations~\cite{mcallester2020formal,song2019understanding}. Moreover,~\cite{tschannen2019mutual,tian2020makes} show that higher mutual information learned by the network does not warrant better representation quality. In our work, we find adding mutual information bits between two views which are irrelevant to downstream tasks can be harmful for the quality of learned representations. Data augmentation plays an important role at favoring certain bits of mutual information than others.

There is a growing number of recent work on the topic of understanding contrastive learning, both theoretically~\cite{arora2019theoretical,tsai2020demystifying,tosh2020contrastive,tian2020understanding,lee2020predicting} and empirically~\cite{wang2020understanding,tian2020makes,zhao2020makes,purushwalkam2020demystifying}. However, little work has been done to study the phenomenon of feature suppression. To our knowledge, we are the first one to quantitatively and systematically study this problem. We believe this is still a very open question and could benefit from more future investigation. Finally, the feature suppression effect in unsupervised contrastive learning that we study in this work may also exist in standard supervised learning (``contrastive loss'' between examples and class labels), as suggested by~\cite{hermann2020shapes,shah2020pitfalls}, though the specific form would be different.
 \section{Conclusion}

In this work, we study three intriguing properties of contrastive losses. In particular, our results highlight that feature suppression is still an open challenge in contrastive learning. While there is a plethora of work on improving contrastive learning, few of them directly aim to address feature suppression. This limitation of contrastive learning becomes a bottleneck for scenarios where existing augmentation cannot fully address the feature suppression phenomenon, and learning would saturate at a level of dissatisfaction.

We would also like to point out some limitations of our study. Firstly, we focus mostly on contrastive learning with explicit negatives (e.g. SimCLR and MoCo). We believe other methods based on clustering and/or without negative pairs would exhibit similar phenomenon but we leave that as future work. Secondly, many of our proposed image datasets are not fully realistic despite being composed from some (challenging) natural image datasets such as ImageNet. We admit it is very hard to explore competing features or multiple objects in a controllable fashion on realistic large scale image datasets.

\section*{Acknowledgements}
We specially thank Geoffrey Hinton for many inspiring discussions and helpful advice. We would also like to thank David Fleet, Simon Kornblith, Mohammad Norouzi, Kevin Swersky and Katherine Hermann for insightful discussions. In addition, we are thankful to William Chan and Sara Sabour for ideas on implementation of sorting on TPUs. We also thank anonymous reviewers for their constructive feedback.

\newpage
{\small
\bibliography{content/ref}
\bibliographystyle{unsrtnat}}

\counterwithin{figure}{section}
\counterwithin{table}{section}
\newpage
\appendix
\section{More on generalized contrastive loss}
\label{app:gloss}

\subsection{Experimental setup}
\label{app:gloss_setup}
 
We follow~\cite{chen2020simple,chen2020big} for the use of augmentations and architectures. By default, we use ResNet-50~\cite{he2016deep} and a 2-layer projection head~\cite{chen2020simple,chen2020big} after the ResNet's average pooling layer. We set the output ($\bm z$) dimensionality to 64 for CIFAR10 and 128 for ImageNet, since increasing them has little effect on the performance. We use a square root learning rate scaling with batch size with a LARS optimizer~\cite{you2017large}, i.e., $\mathrm{LearningRate}=0.075\times\sqrt\mathrm{BatchSize}$ for ImageNet and $\mathrm{LearningRate}=0.2\times\sqrt\mathrm{BatchSize}$ for CIFAR-10. The batch size and training epoch will be specified for each experiment. We use the linear evaluation protocol, i.e. the accuracy of a trained linear classifier on the learned features is used as a proxy for representation quality.

When comparing the standard contrastive loss (i.e. NT-Xent in Eq.~\ref{eq:nt_xent}) and other instantiations of the generalized contrastive loss (in Table~\ref{tab:gloss_comp}), we optimize the hyper-parameters for different losses (for NT-Xent loss, we set $\tau=0.2$; for decoupled NT-Xent loss, we set $\tau=1.0,\lambda=0.1$; for SWD based losses, we set $\lambda=5$
; and since we use mean squared error instead of $\ell_2$ distance in alignment loss for losses in Table~\ref{tab:gloss_comp}, we find it helpful to scale the loss by 1000 when the hidden vector $\bm z$ is normalized). A batch size of 128 is used for CIFAR-10, and 1024 is used for ImageNet.

\subsection{Temperature $\tau$ is (within a range) inversely correlated to weighting $\lambda$ of distribution loss}
\label{app:tau_lambda}

\paragraph{Both temperature $\tau$ and weighting $\lambda$ control how well the representations fit the prior.} To see how well the learned distribution matches the prior distribution (e.g. Gaussian), we randomly project the  (high-dimensional) representation vectors into 1-D space and plot the histogram distribution. For prior distribution of Gaussian or uniform in hypersphere, these random projections in 1-D space should be Gaussian like.

\begin{figure*}[h]
    \centering\small
    \begin{subfigure}{.45\textwidth}
      \centering
    \includegraphics[trim=0 0 330 0,clip,width=0.99\textwidth]{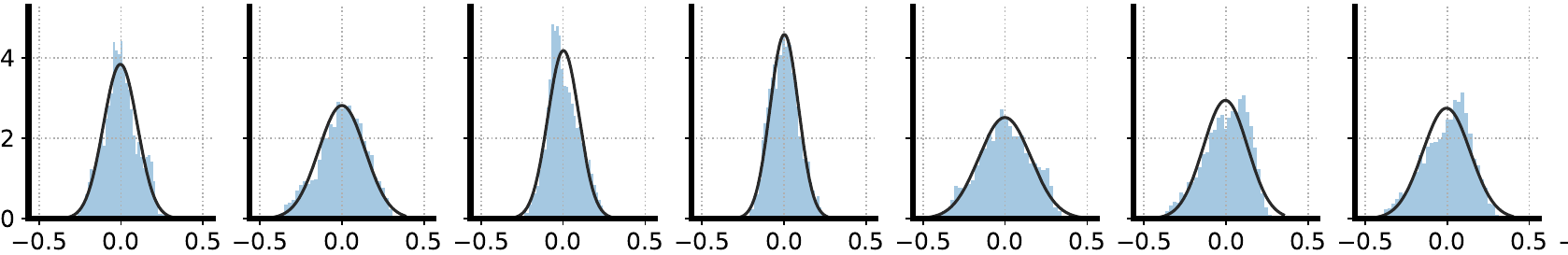}
      \caption{SWD ($\lambda=0.5$)}
    \end{subfigure}
    ~~~~~~~~~~~~
    \begin{subfigure}{.45\textwidth}
      \centering
    \includegraphics[trim=0 0 330 0,clip,width=0.99\textwidth]{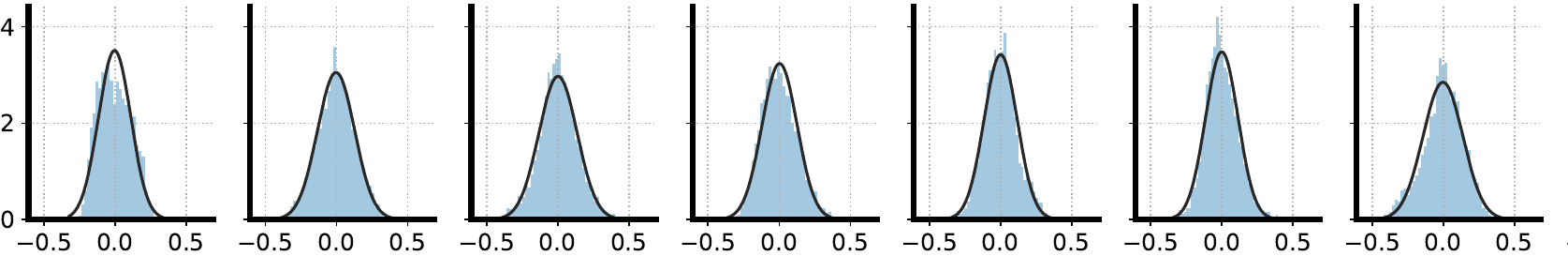}
      \caption{NT-XENT ($\tau=0.4$)}
    \end{subfigure}
    \\
    \begin{subfigure}{.45\textwidth}
      \centering
    \includegraphics[trim=0 0 330 0,clip,width=0.99\textwidth]{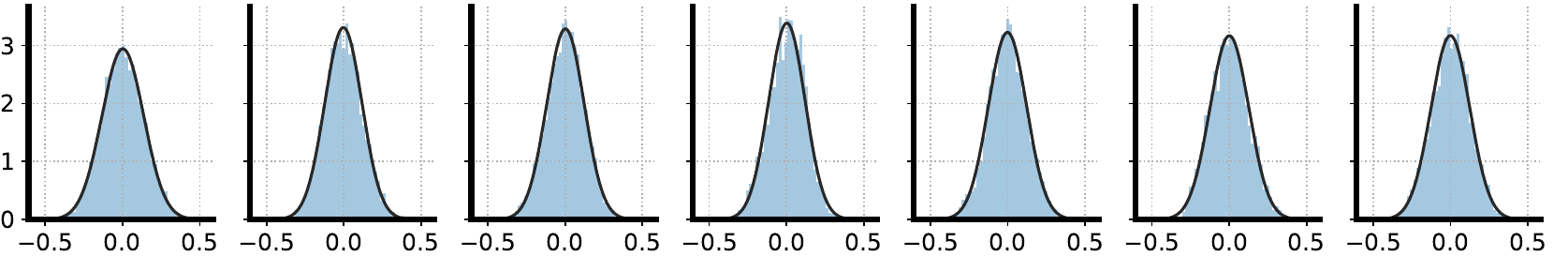}
      \caption{SWD ($\lambda=5$)}
    \end{subfigure}
    ~~~~~~~~~~~~
    \begin{subfigure}{.45\textwidth}
      \centering
    \includegraphics[trim=0 0 330 0,clip,width=0.99\textwidth]{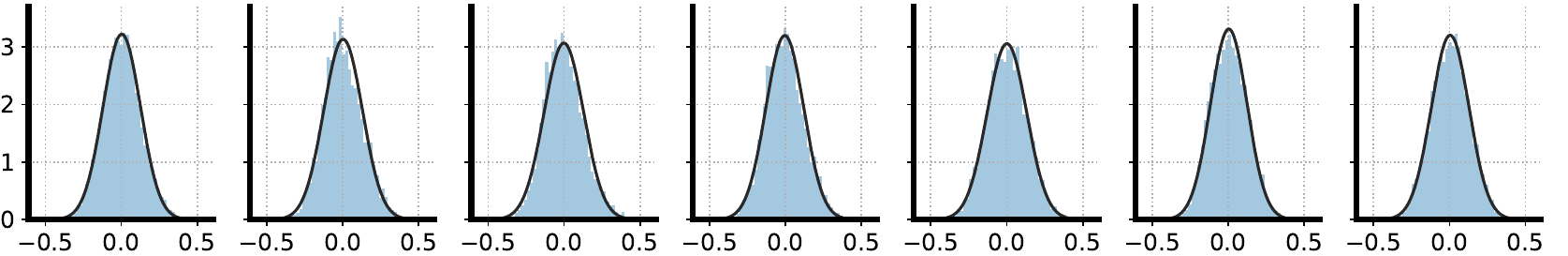}
      \caption{NT-XENT ($\tau=0.2$)}
    \end{subfigure}
    \\
    \begin{subfigure}{.45\textwidth}
      \centering
    \includegraphics[trim=0 0 330 0,clip,width=0.99\textwidth]{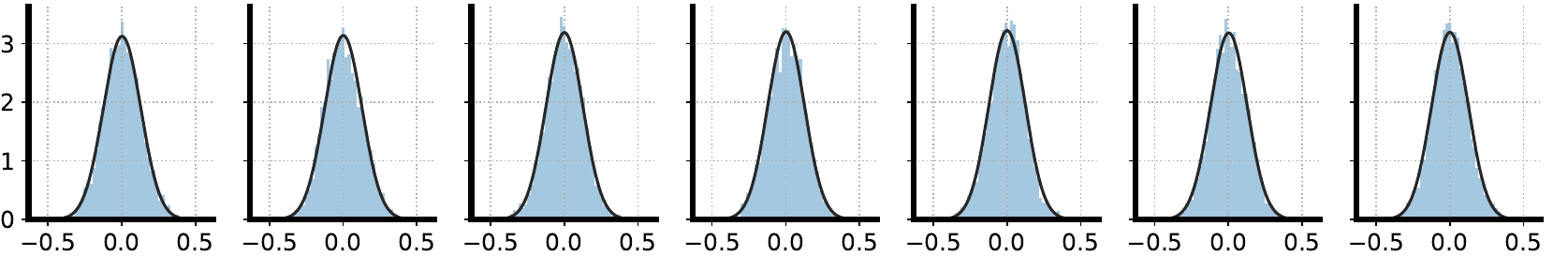}
      \caption{SWD ($\lambda=50$)}
    \end{subfigure}
    ~~~~~~~~~~~~
    \begin{subfigure}{.45\textwidth}
      \centering
    \includegraphics[trim=0 0 330 0,clip,width=0.99\textwidth]{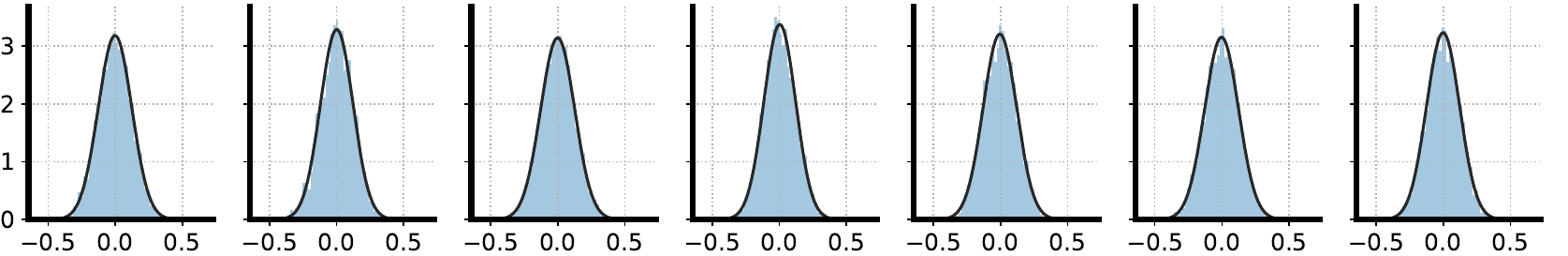}
      \caption{NT-XENT ($\tau=0.1$)}
    \end{subfigure}
    \caption{\label{fig:proj_dist_swd} Distribution of random orthogonal projection of output vectors on CIFAR-10 test set (each small plot has its own random projection direction). For SWD (uniform hypersphere) loss, distribution becomes more Gaussian as $\lambda$ increases. For NT-Xent loss, the distribution becomes more Gaussian as $\tau$ decreases.}
\end{figure*}

Figure \ref{fig:proj_dist_swd} shows random orthogonal projection of representation from CIFAR-10 test set. We see that both weighting ($\lambda$ in Eq.~\ref{eq:gc}) and the temperature scaling ($\tau$ in Eq.~\ref{eq:nt_xent}) have the effect of controlling distribution matching term, but they have an inverse correlation.
In other words, using a higher temperature has similar effect as setting a larger weighting of distribution matching term.

\paragraph{Decoupled NT-Xent loss.} It is worth noting that temperature $\tau$ in the rewritten NT-Xent loss (Eq.~\ref{eq:nt_xent_scaled}) appears in two places, one as the scaling of the distribution loss term, and the other as the width of Gaussian kernel. They do not necessarily need to be the same, so we could decouple them as follows.

\begin{equation}
\small
\label{eq:nt_xent_decoupled}
\begin{aligned}
    \mathcal{L}^{\text{Decoupled }\mathrm{NT}\text{-}\mathrm{Xent}} =  -\frac{1}{n}\sum_{i,j}\mathrm{sim}(\bm z_i, \bm z_j)
    + \lambda \frac{1}{n}\sum_i\log\sum_{k=1}^{2n} \one{k \neq i}\exp(\mathrm{sim}(\bm z_i, \bm z_k)/\tau)
\end{aligned}
\end{equation}

\begin{figure}[h]
    \centering
    \includegraphics[width=0.85\textwidth]{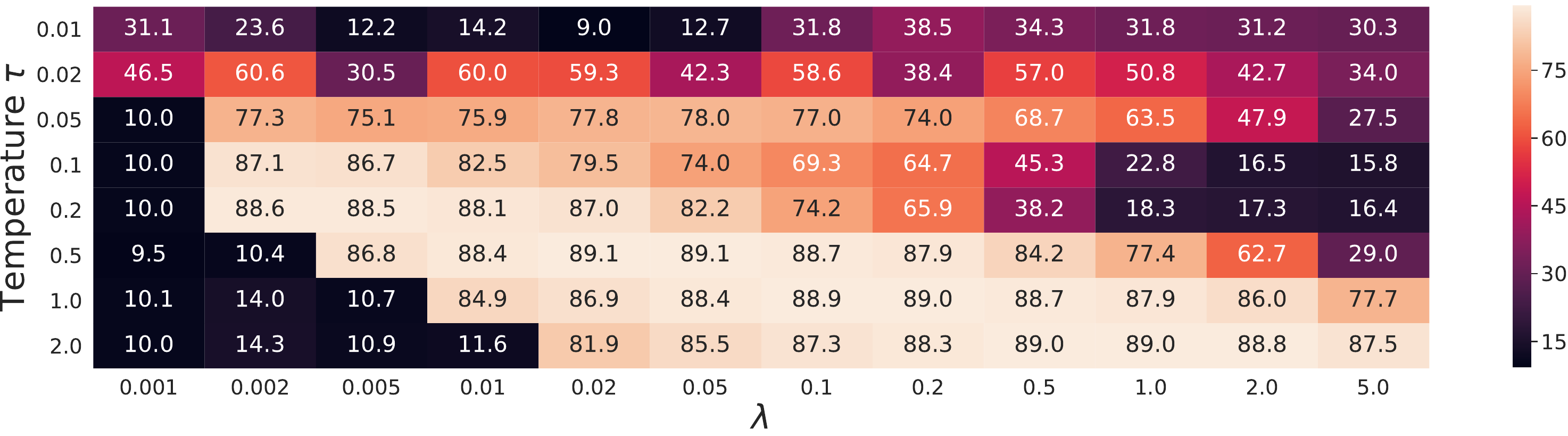}
    \caption{\label{fig:linear_temp_weight}Linear evaluation of ResNet-18 trained on CIFAR-10 (200 epochs) using decoupled NT-Xent loss (Eq.~\ref{eq:nt_xent_decoupled}). The temperature $\tau$ and the weighting $\lambda$ are mostly inverse correlated.}
\end{figure}

The decoupling allows us to study the effects of them separately. So we tune $\tau$ and $\lambda$ separately for the decoupled NT-xent loss. Figure~\ref{fig:linear_temp_weight} shows the linear evaluation of ResNet-18 trained in 200 epochs. We see that the temperature $\tau$ and the weighting $\lambda$ are inversely correlated for most range. In practice one could simply fix one and tune the other.

\subsection{Linear evaluation of generalized contrastive losses on CIFAR-10 and ImageNet}
\label{app:linear_eval}

Table~\ref{tab:linear_eval_num_cifar10},~\ref{tab:linear_eval_num_imagenet_h2} and \ref{tab:linear_eval_num_imagenet_h3} show linear evaluation performance of ResNet-50 trained with different losses (numerical results of Figure~\ref{fig:comploss_linear}). Similar to~\cite{chen2020simple,chen2020big}, a square root learning rate is used. In addition, results of different batch sizes are also compared, and we find the differences are small with reasonable sizes (e.g. 128 for CIFAR-10 and 1024 for ImageNet).

\begin{table}[h]
    \centering
    \footnotesize
    \caption{\label{tab:linear_eval_num_cifar10}Linear evaluation accuracy (top-1) of ResNet-50 trained with different losses on CIFAR-10.}
    \begin{tabular}{llrrrr}
    \toprule
                              & Epoch &   100 &   200 &   400 &   800 \\
    Loss & Batch size &       &       &       &       \\
    \midrule
    \multirow{4}{*}{NT-Xent} & 128  &  87.4 &  91.0 &  93.0 &  93.9 \\
                              & 256  &  88.0 &  91.3 &  93.0 &  93.6 \\
                              & 512  &  87.9 &  91.3 &  92.9 &  93.7 \\
                              & 1024 &  88.2 &  91.2 &  92.7 &  93.3 \\
\midrule
    \multirow{4}{*}{Decoupled NT-Xent} & 128  &  87.8 &  91.0 &  93.0 &  94.0 \\
                              & 256  &  87.7 &  91.1 &  92.8 &  93.6 \\
                              & 512  &  87.5 &  91.3 &  92.7 &  93.6 \\
                              & 1024 &  87.5 &  91.0 &  92.6 &  93.7 \\
\midrule
    \multirow{4}{*}{SWD (normal)} & 128  &  86.3 &  90.5 &  92.8 &  93.8 \\
                              & 256  &  86.2 &  90.8 &  93.1 &  94.1 \\
                              & 512  &  85.0 &  90.7 &  92.9 &  94.1 \\
                              & 1024 &  83.3 &  89.9 &  93.0 &  93.9 \\
\midrule
    \multirow{4}{*}{SWD (uniform hypercube)} & 128  &  85.1 &  90.1 &  92.6 &  93.4 \\
                              & 256  &  84.6 &  89.9 &  92.9 &  93.8 \\
                              & 512  &  83.1 &  89.8 &  92.8 &  93.8 \\
                              & 1024 &  81.3 &  88.3 &  92.2 &  93.6 \\
\midrule
    \multirow{4}{*}{SWD (uniform hypersphere)} & 128  &  87.0 &  90.9 &  92.9 &  93.8 \\
                              & 256  &  87.1 &  90.9 &  92.5 &  93.7 \\
                              & 512  &  86.6 &  90.8 &  92.9 &  93.4 \\
                              & 1024 &  86.0 &  90.3 &  92.5 &  93.2 \\
    \bottomrule
    \end{tabular}
\end{table}

\begin{table}[h]
    \centering
    \footnotesize
    \caption{\label{tab:linear_eval_num_imagenet_h2}Linear evaluation accuracy (top-1) of ResNet-50 trained with different losses on ImageNet (with 2-layer projection head).}
    \begin{tabular}{llrrrr}
    \toprule
                              & Epoch &   100 &   200 &   400 &   800 \\
    Loss & Batch size &       &       &       &       \\
    \midrule
    \multirow{3}{*}{NT-Xent} & 512  &  65.4 &  67.3 &  68.7 &  69.3 \\
                              & 1024 &  65.6 &  67.6 &  68.8 &  69.8 \\
                              & 2048 &  65.3 &  67.6 &  69.0 &  70.1 \\
\midrule
    \multirow{3}{*}{Decoupled NT-Xent} & 512  &  65.8 &  67.6 &  68.9 &  69.5 \\
                              & 1024 &  66.0 &  67.9 &  69.0 &  70.1 \\
                              & 2048 &  65.8 &  67.9 &  69.3 &  70.2 \\
\midrule
    \multirow{3}{*}{SWD (normal)} & 512  &  64.9 &  66.8 &  68.0 &  69.0 \\
                              & 1024 &  65.0 &  67.1 &  68.2 &  69.3 \\
                              & 2048 &  65.0 &  66.9 &  68.4 &  69.7 \\
\midrule
    \multirow{3}{*}{SWD (uniform hypercube)} & 512  &  64.3 &  66.4 &  67.8 &  68.7 \\
                              & 1024 &  64.2 &  66.5 &  67.9 &  68.9 \\
                              & 2048 &  63.9 &  66.6 &  67.9 &  69.0 \\
\midrule
    \multirow{3}{*}{SWD (uniform hypersphere)} & 512  &  65.6 &  67.7 &  69.0 &  70.0 \\
                              & 1024 &  65.8 &  67.9 &  69.0 &  69.6 \\
                              & 2048 &  65.6 &  67.8 &  69.2 &  69.8 \\
    \bottomrule
    \end{tabular}
\end{table}

\begin{table}[h]
    \centering
    \footnotesize
    \caption{\label{tab:linear_eval_num_imagenet_h3}Linear evaluation accuracy (top-1) of ResNet-50 trained with different losses on ImageNet (with 3-layer projection head).}
    \begin{tabular}{llrrrr}
    \toprule
                              & Epoch &   100 &   200 &   400 &   800 \\
    Loss & Batch size &       &       &       &       \\
    \midrule
    \multirow{3}{*}{NT-Xent} & 512  &  66.6 &  68.4 &  70.0 &  71.0 \\
                              & 1024 &  66.8 &  68.9 &  70.1 &  70.9 \\
                              & 2048 &  66.8 &  69.1 &  70.4 &  71.3 \\
\midrule
    \multirow{3}{*}{Decoupled NT-Xent} & 512  &  66.8 &  68.4 &  69.6 &  70.6 \\
                              & 1024 &  66.6 &  68.9 &  69.9 &  70.8 \\
                              & 2048 &  66.6 &  69.0 &  70.1 &  70.8 \\
\midrule
    \multirow{3}{*}{SWD (normal)} & 512  &  66.5 &  68.4 &  69.8 &  70.8 \\
                              & 1024 &  66.6 &  68.8 &  70.1 &  71.1 \\
                              & 2048 &  66.7 &  69.1 &  70.2 &  71.1 \\
\midrule
    \multirow{3}{*}{SWD (uniform hypercube)} & 512  &  66.1 &  68.3 &  69.7 &  70.7 \\
                              & 1024 &  66.3 &  68.5 &  70.0 &  71.3 \\
                              & 2048 &  65.8 &  68.2 &  70.1 &  71.1 \\
\midrule
    \multirow{3}{*}{SWD (uniform hypersphere)} & 512  &  66.5 &  68.3 &  69.5 &  70.5 \\
                              & 1024 &  66.6 &  68.6 &  69.8 &  70.8 \\
                              & 2048 &  66.5 &  68.7 &  70.2 &  70.9 \\
    \bottomrule
    \end{tabular}
\end{table}

\section{More on feature suppression}
\label{app:fsupp}

\subsection{Extra results on CIFAR-10 and ImageNet with random bits added}
\label{app:ext_probing_result_bit}

\begin{figure}[H]
    \centering
    \begin{subfigure}{.9\textwidth}
      \centering
      \includegraphics[width=0.98\linewidth]{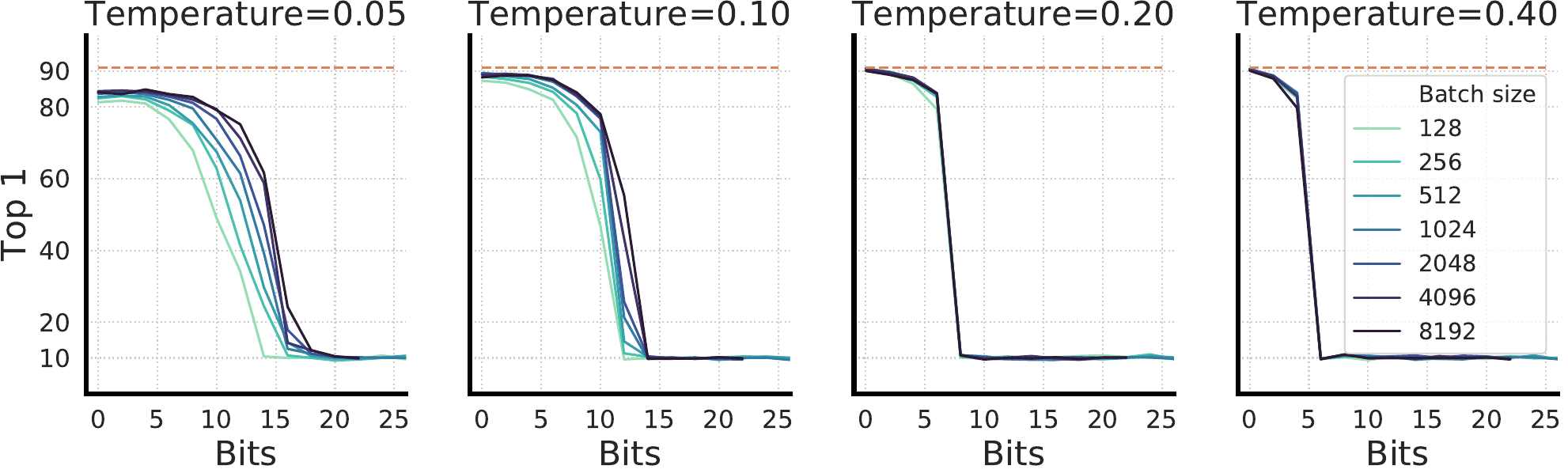}
      \caption{Standard NT-Xent}
    \end{subfigure}
    ~~\\~~\\\begin{subfigure}{.9\textwidth}
      \centering
      \includegraphics[width=0.98\linewidth]{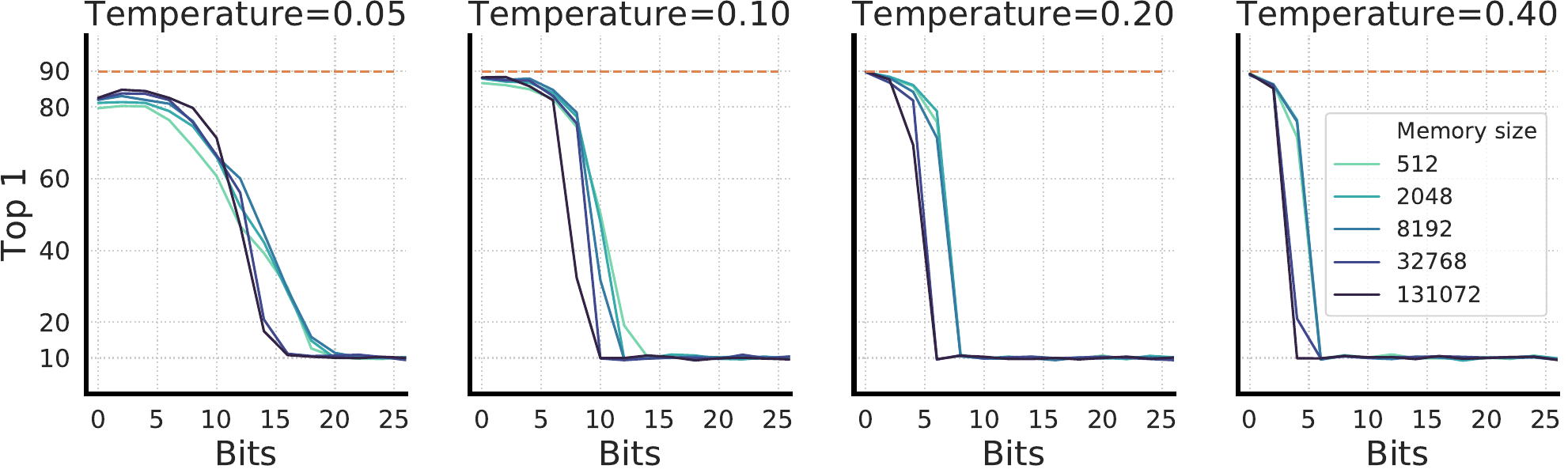}
      \caption{NT-Xent with Momentum Contrast (MoCo)}
    \end{subfigure}
    ~~\\~~\\\begin{subfigure}{.9\textwidth}
      \centering
      \includegraphics[width=0.98\linewidth]{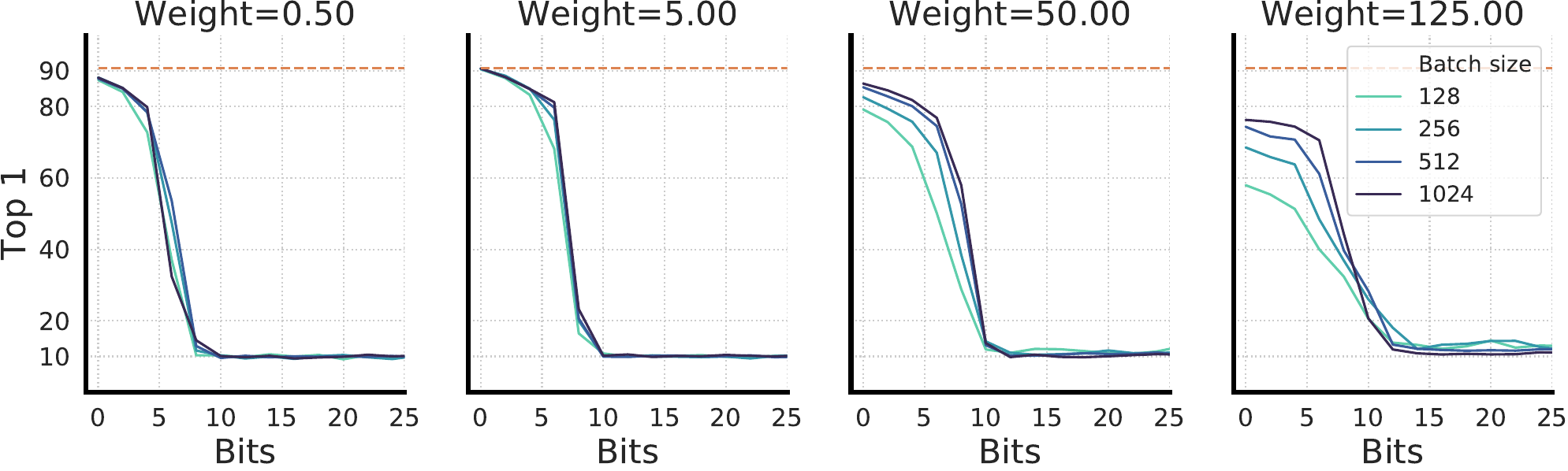}
      \caption{SWD (uniform hypersphere)}
    \end{subfigure}
    \caption{\label{fig:randb_cifar10_ext}Linear evaluation accuracy on CIFAR-10 of ResNet-18 (400 epochs) when different random bits are added. Different contrastive losses and batch sizes are compared.}
\end{figure}

Figure~\ref{fig:randb_cifar10_ext} shows linear evaluation on CIFAR-10 with different random bits added trained with a wider range of batch sizes. It is worth noting that the bits (in the x-axis) are calculated based on the total size of uniform integer distribution. However, this is an overestimation of actual bits as due to collision in generated integers. 
We observe that the linear evaluation accuracy decreases quickly with a few bits of the extra channel competing feature added. And this detrimental effect on the representation quality cannot be avoided by different contrastive loss functions, batch sizes, or memory mechanism in momentum contrast~\cite{he2019momentum}. Although a smaller temperature ($\tau$) or larger weighting ($\lambda$) slightly mitigate the degeneration effect, its baseline performance when no extra bits are added is also worse. {With less than 15 bits of competing features added, the representation quality degenerates to the level where RGB channels are completely ignored.} 

\begin{figure}[H]
    \centering
    \begin{subfigure}{.27\textwidth}
      \centering
      \includegraphics[width=0.98\linewidth]{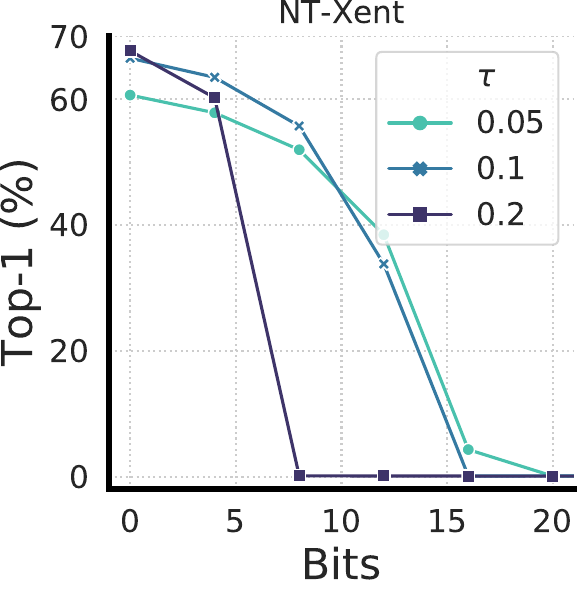}
\caption*{}
    \end{subfigure}
    \begin{subfigure}{.27\textwidth}
      \centering
      \includegraphics[width=0.98\linewidth]{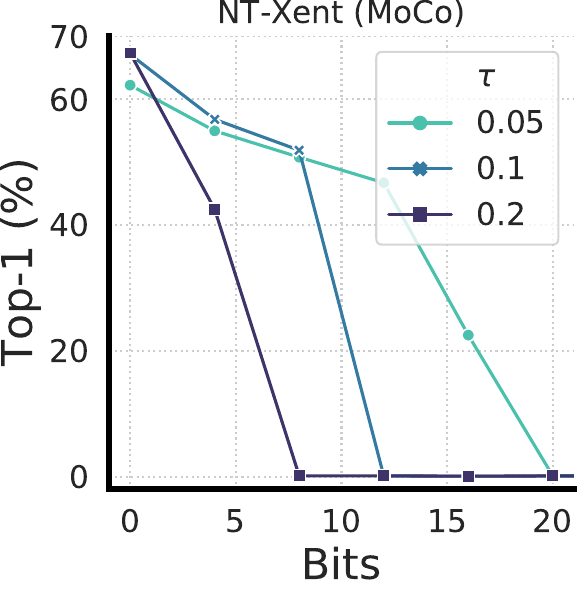}
\caption{\label{fig:linear_randb_imnet}ImageNet.}
    \end{subfigure}
    \begin{subfigure}{.27\textwidth}
      \centering
      \includegraphics[width=0.98\linewidth]{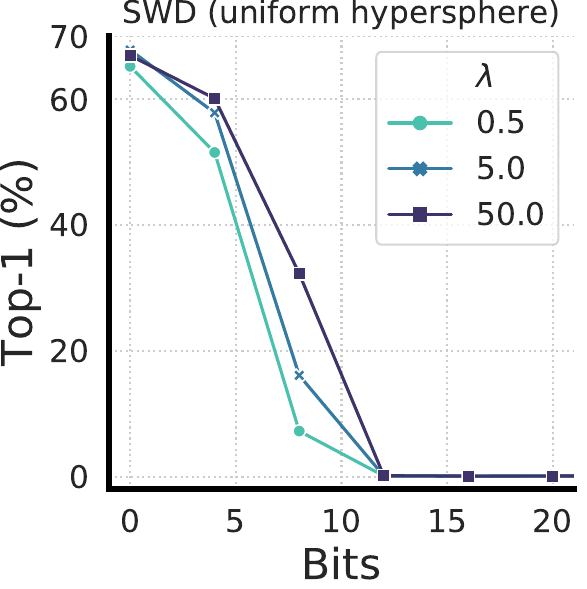}
\caption*{}
    \end{subfigure}
    \caption{\label{fig:linear_randb_imagenet} Linear evaluation of learned features when a few bits of competing features added on ImageNet. A few bits added completely disable contrastive learning (across various batch size or losses).
}
\end{figure}

Similar results are shown for ImageNet as shown in Figure~\ref{fig:linear_randb_imagenet}.

\newpage
\subsection{Distribution matching loss, $\mathrm{LogSumExp}$ or SWD, saturates with a few bits of entropy}
\label{app:dist_sat}

Here we study the saturation of distribution matching loss (based on $\mathrm{LogSumExp}$ or SWD), without presence of the alignment term. To do so, we create square images with $k$ binary channels (instead of RGB channels), and all pixels at different locations of a $32\times 32$ image share the same value, this allows us to use the same architecture as one for CIFAR-10 (i.e. ResNet-18 and 2-layer projection head with output dimensionality of 64). We note that this experiment can also be conducted on images of $1\times 1$ size with other architecture. It is not difficult to see the entropy of this dataset is $k$ bits. A mini-batch of data points (without augmentations) are first encoded via the network, and then the distribution matching loss is defined on the network's outputs. The network is trained for 400 epochs, and longer training epochs makes little difference. 

\begin{figure}[h]
    \begin{subfigure}{.85\textwidth}
      \centering
      \includegraphics[width=0.98\linewidth]{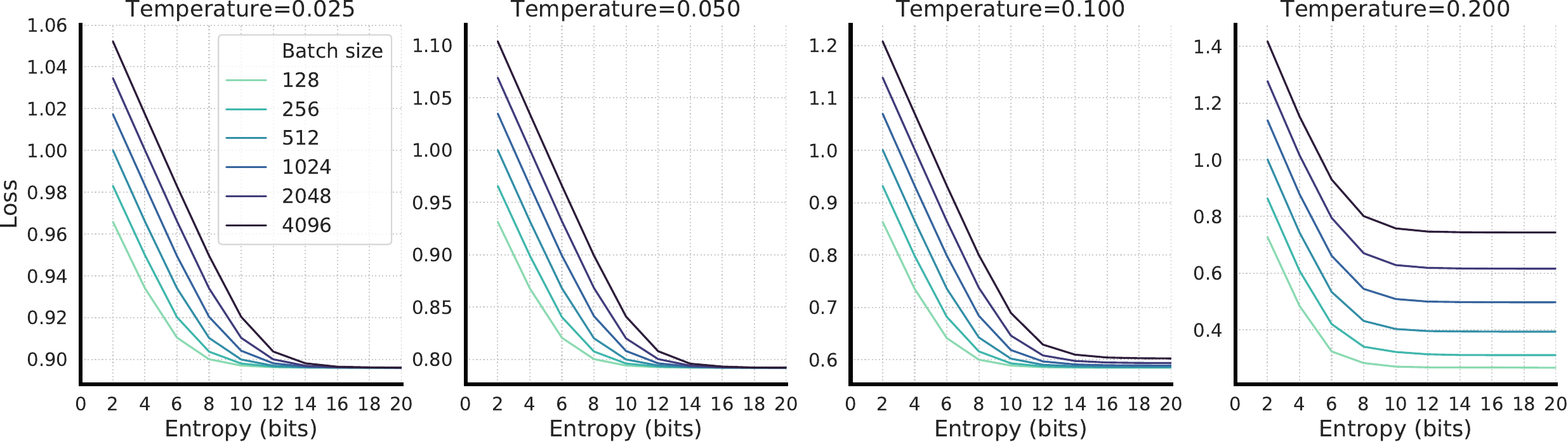}
      \caption{$\mathrm{LogSumExp}$.}
    \end{subfigure}
    \begin{subfigure}{.65\textwidth}
      \centering
      \includegraphics[width=0.98\linewidth]{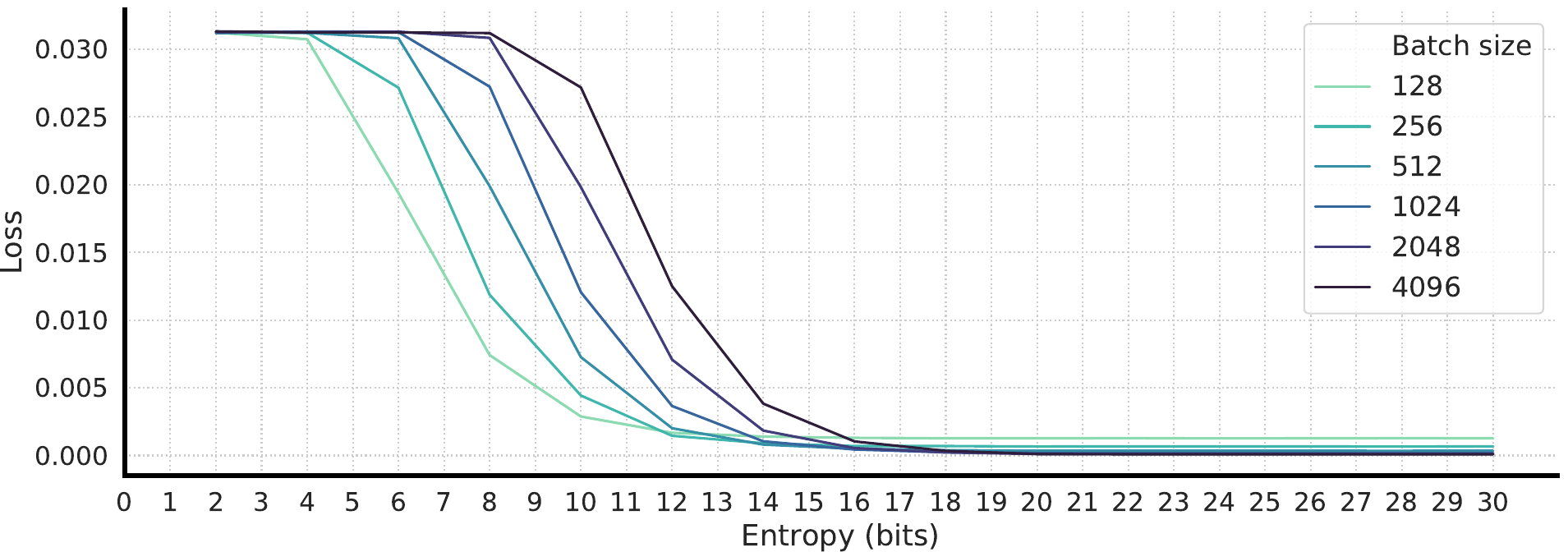}
      \caption{SWD (uniform hypersphere).}
    \end{subfigure}
    \centering \caption{\label{fig:distm} Distribution matching loss saturates quickly with a few bits of entropy. The saturation varies slightly across batch sizes.}
\end{figure}

Figure~\ref{fig:distm} shows that distribution loss saturates quickly with a few bits of entropy in the dataset (same or less bits in representations), and both temperature and batch sizes have effects on the saturation behavior. It also shows that linear increase of bits in representation requires exponentially increase of batch size, which is not sustainable as the required batch size can quickly go beyond the size of the dataset (e.g., 30 bits would require more than 1 billion batch size, which is larger than most of the existing datasets). This is one of the main reasons why data augmentation is critical for contrastive learning - that the network can learn a few bits that give rise to useful representations.

\vfill 
\end{document}